\documentclass[letterpaper, 10 pt, conference]{ieeeconf}  

\usepackage{times}
\usepackage{epsfig}
\usepackage{graphicx}
\usepackage{amsmath}
\usepackage{amssymb}
\usepackage{multirow}
\usepackage{caption,subcaption}
\usepackage{floatrow}
\usepackage[dvipsnames]{xcolor}
\usepackage{cite}

\usepackage{color}
\usepackage{soul}

\DeclareMathOperator*{\argmin}{argmin}

\definecolor{myRed}{rgb}{0.8, .2, .2}

\definecolor{myBlue}{rgb}{0.2, .2, .8}

\IEEEoverridecommandlockouts                              

\title{\LARGE \bf
AR Mapping: Accurate and Efficient Mapping for Augmented Reality
}

\author{Rui Huang$^{1}$, Chuan Fang, Kejie Qiu, Le Cui, Zilong Dong, Siyu Zhu, Ping Tan
 \thanks{$^{1}$ All the authors are with Alibaba Group, Hangzhou, China.
  {\tt\small rui.hr@alibaba-inc.com}}%
}

\begin{document}

\maketitle
\thispagestyle{empty}
\pagestyle{empty}

\begin{abstract}
Augmented reality (AR) has gained increasingly attention from both research and industry communities. By overlaying digital information and content onto the physical world, AR enables users to experience the world in a more informative and efficient manner. As a major building block for AR systems, localization aims at determining the device's pose from a pre-built ``map'' consisting of visual and depth information in a known environment. While the localization problem has been widely studied in the literature, the ``map'' for AR systems is rarely discussed. In this paper, we introduce the AR Map for a specific scene to be composed of 1) color images with 6-DOF poses; 2) dense depth maps for each image and 3) a complete point cloud map. We then propose an efficient end-to-end solution to generating and evaluating AR Maps. Firstly, for efficient data capture, a backpack scanning device is presented with a unified calibration pipeline. Secondly, we propose an AR mapping pipeline which takes the input from the scanning device and produces accurate AR Maps. Finally, we present an approach to evaluating the accuracy of AR Maps with the help of the highly accurate reconstruction result from a high-end laser scanner. To the best of our knowledge, it is the first time to present an end-to-end solution to efficient and accurate mapping for AR applications. 

\end{abstract}

\section{Introduction}
Augmented reality (AR) has gained increasingly attention from both research and industry communities. Many AR applications have emerged in areas of entertainment, construction and indoor navigation. By overlaying digital information and content onto the physical world, AR enables users to experience the world in a more informative and efficient manner. 
According to Bimber et.al,~\cite{bimber2005spatial}, AR systems are built upon three major blocks including tracking and registration, display technology and real time rendering. In order to render the virtual object correctly, the system needs to constantly determine the pose within the environment of the user with respect to the virtual object. Those applications such as Google Map AR~\cite{googleAR} and Pokemon Go~\cite{pokemongo} start with re-localizing the device's absolute pose outdoor with the help of GPS, and then track the device's subsequent motion continuously using motion tracking systems such as ARCore~\cite{ARcore}. 

It has been shown that localization is a major building block for AR systems~\cite{bimber2005spatial,mekni2014augmented}. Currently, the state-of-the-art approaches to localization in arbitrary environments can be divided into two categories: 1) sparse feature matching~\cite{sarlin2020superglue}; 2) scene coordinate map regression~\cite{brachmann2017dsac}. Both categories needs a pre-built ``map'' including color images and corresponding sparse or dense depth information. However, while the localization problem has been widely studied in the literature, the ``map'' for AR systems has not been well defined so that we propose a framework for building and evaluating AR Maps.


\begin{figure}  
\includegraphics[width=\linewidth]{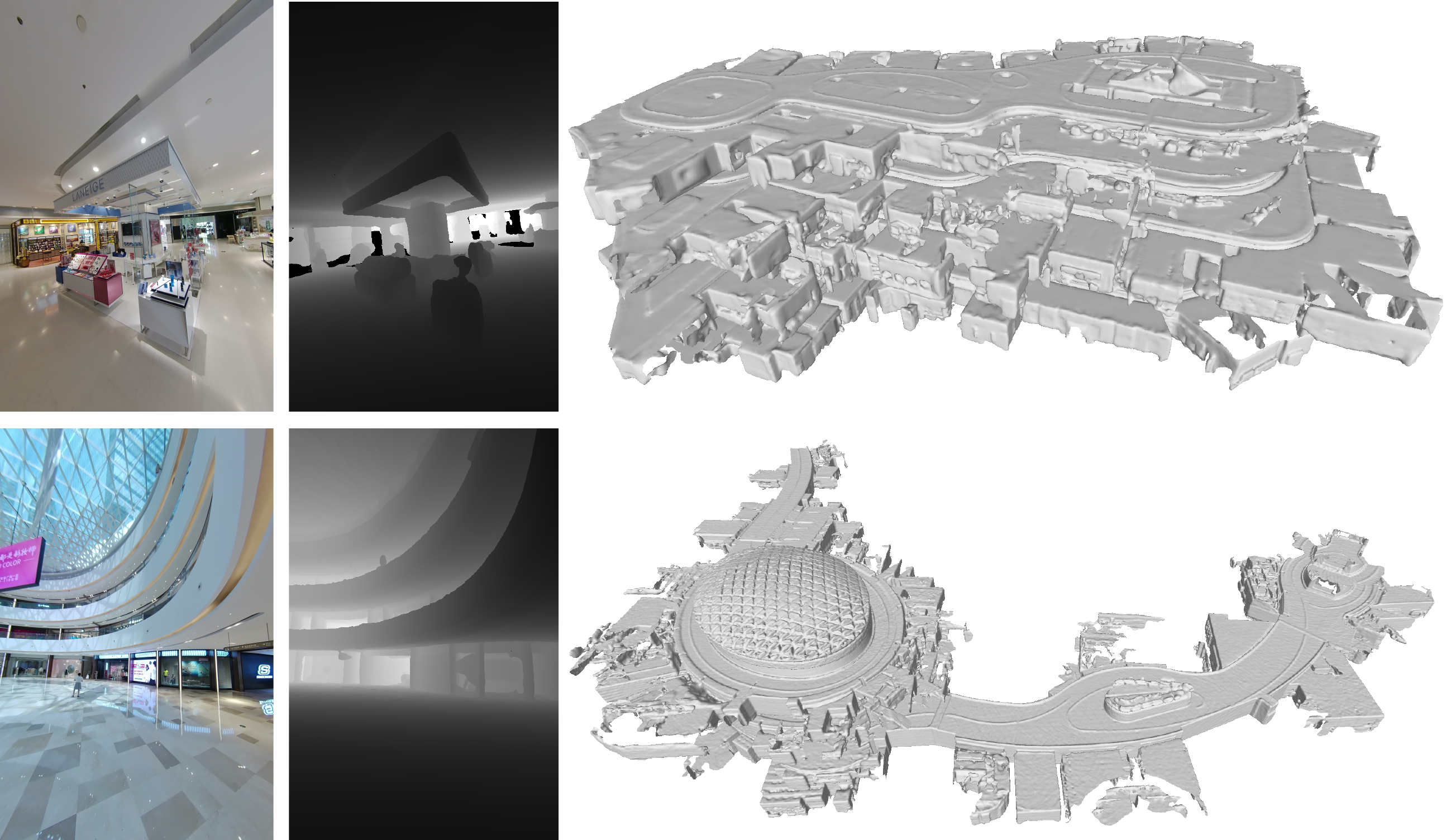}
\caption{The AR Maps of two large scenes: WestCity (1st row) and EFC Mall (2nd row). From left to right, it shows the captured color image, corresponding depth map and reconstructed 3D model from point cloud map.}
\label{fig:armap_content}
\end{figure}

We introduce AR Maps which can be used in any AR systems. For a specific scene, an AR Map is composed of 1) color images with 6-DOF poses; 2) dense depth map for each image and 3) a complete point cloud map. An AR Map is the set of information which enables basic functions such as localization and occlusion detection in AR systems. To ensure that the AR Map meets the requirements of AR applications, the following issues need to be considered.

\begin{itemize}
  \item How to capture the visual and depth information of a scene efficiently so that the AR Map can be updated frequently with low cost.
  \item How a mapping system can be designed to process the raw data and generate an accurate AR Map. The accuracy of depth and geometry in local areas is critical for some tasks such as occlusion detection and realistic rendering in AR systems. 
  \item How to evaluate an AR Map to ensure that its accuracy meets the requirements of AR applications.
\end{itemize}


To address these issues, we propose an end-to-end solution to generating and evaluating AR Maps. Firstly, we present a scanning device in the form of backpack equipped with 2 multi-beam lidars, a panoramic camera with 4 fisheye lens and an inertial measurement unit (IMU). Lidars are active sensors which directly measure distance by emitting laser light to the target. In constrast to RGB cameras, they are not affected by the illumination condition or texture richness of the environment. We capture dense depth by multi-beam lidars and obtain color images by the panoramic camera. To calibrate a device with multi-model sensors, it usually requires overlap between all sensors and time-consuming manual placement of calibration board~\cite{lee2020unified}. We further propose an unified extrinsic calibration approach which requires only one-time installment and data capture. The subsequent calibration process is fully automated. 

Secondly, we propose a lidar-based AR mapping system to construct accurate AR Maps. Traditional lidar odometry and mapping systems~\cite{zhang2014loam, zhang2018laser} focus on motion estimation. The mapping process is usually done by transforming the point cloud to a global frame. To ensure efficient and accurate mapping, we firstly introduce a lidar odometry module with some improvements based on the classic LOAM~\cite{zhang2014loam} system. These improvements include an enhanced feature selection model and filtering strategy to remove outliers. To correct the accumulated error in odometry for large-scale scenes, we further propose a submap-based global optimization module which optimizes the global trajectory using constraints of loop closing and consistency between neighboring submaps. This module eliminates the drifting error while maintaining the local map consistency. Moreover, the final point cloud map is generated from a stable map fusion module. By enforcing the map points with sufficient observations, this module guarantees a clean map even in highly dynamic environments. Once obtaining the optimized global trajectory and point cloud map, the camera poses for color images can be interpolated from the lidar poses and the corresponding depth maps are rendered from the reconstructed 3D mesh. The content of resulting AR Maps are demonstrated in Fig.~\ref{fig:armap_content}.

Finally, we give an approach to evaluating the accuracy of AR Maps based on the highly accurate point cloud reconstruction from a high-end laser scanner. The accuracy of dense depth maps and 6-DOF poses of color images are also evaluated according to their local consistency. The experiments show that our system is able to generate accurate AR Maps efficiently. 

\begin{figure*}[ht]
    \centering
    \includegraphics[width=\textwidth]{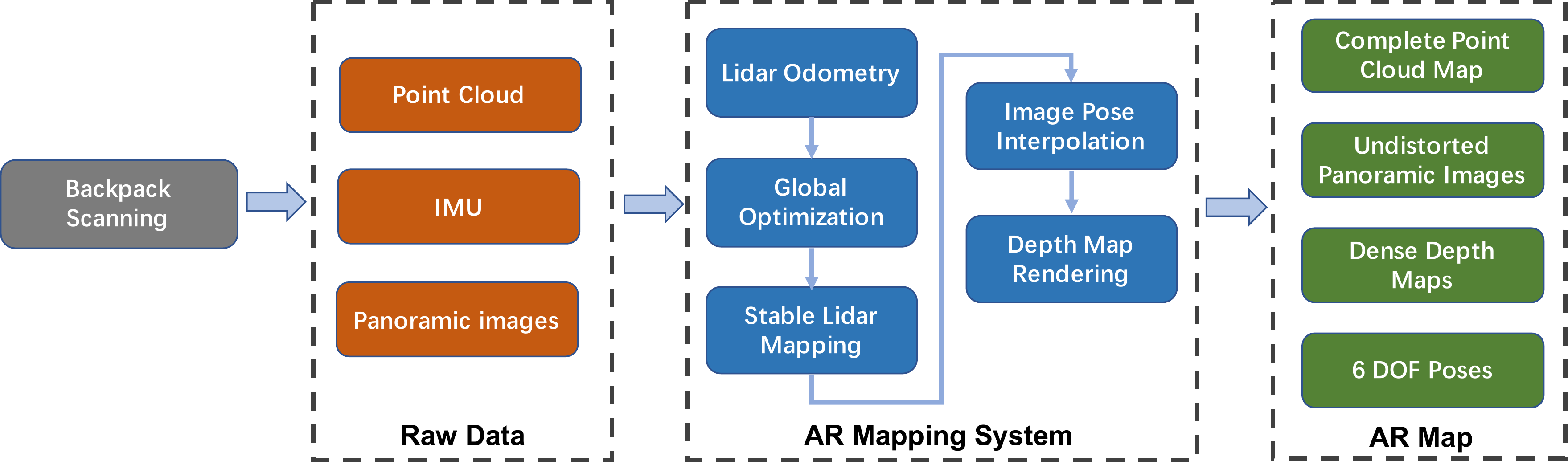}
    \caption{Our end-to-end solution to generating AR Maps with a backpack scanning system and the AR mapping system.}
    \label{fig:armapping_pipeline}
\end{figure*}

\section{Related Works}
\subsection{3D Scanning Devices and Unified Multi-Sensor Calibration}
3D scanning devices are used to capture data for 3D reconstruction and mapping. They are different in terms of the sensor combination and scanning method. Here we only review those devices which can generate the content of an AR Map. Matterport Pro2 3D camera~\cite{matterport} uses a structured light (infrared) sensor to obtain the 3D point cloud. At each scan, the 3D camera is mounted at a tripod and rotates by 360 degrees. It takes 20 seconds for each scan and the maximum operating distance is 4.5 meters. Another 3D scanner, Leica BLK360~\cite{leica360}, uses a rotating laser scanner to capture the point cloud map. Utilizing the same rotating and scanning manner, It takes minutes for each scan and requires manual editing to register the point cloud from all scans into a final 3D map. Both devices are able to generate panoramic images by stitching images captured at different angles during rotation. 

Since those 3D scanners can be fixed on a tripod and rotating is controlled precisely by a motor, they are usually able to produce high-precision point cloud. However, the capturing process is time-consuming and thus infeasible for AR Map generation in large-scale scenes. Instead of capturing at a fixed spot per scan, mobile scanners such as handheld or backpack scanners capture the point cloud data with lidars by moving around in the environment. Leica Pegasus~\cite{leicaBackpack} is a high-cost backpack mobile scanner equipped with 5 color cameras, 2 VLP-lidars and GNSS. It is designed for outdoor large scale mapping. Kaarta Contour is a handheld scanner equipped with a continuously rotating laser scanner and an onboard HD color camera which has too limited FOV to ensure the visual coverage in the scene. We therefore design a backpack scanning system with multi-beam lidars and a panoramic camera for efficient data capture of AR maps.

To calibrate a system with multi-modal sensors, Lee et al.~\cite{lee2020unified} propose an unified multi-camera multi-lidar calibration approach by using only one calibration board. Since the overlap between at least two sensors are required, the checkerboard needs to be manually placed at multiple spots. Jeong et al.~\cite{jeong2019road} present a framework for target-less extrinsic calibration of stereo cameras and lidars. They make use of the static road markings as features. However, this approach is specifically designed for system on a car in city environment, which limits its extension to other general systems. We therefore propose an unified extrinsic calibration method for multi-camera multi-lidar calibration which needs easy deployment of calibration environment and only one-time data capture.

\subsection{Lidar-based Odometry and Mapping}
Lidar Odometry generally estimates the motion by computing the relative pose of consecutive scans. Zhang and Singh proposes the classic lidar odometry and mapping (LOAM) system~\cite{zhang2014loam}. Based on the idea of LOAM, many works have been proposed to improve the accuracy and robustness of lidar odometry and mapping. Lego-LOAM~\cite{shan2018lego} is a light-weight and ground-optimized LOAM system designed for ground vehicle mapping in outdoor. However it assums that the vehicle is always moving on a plane. A loosely-coupled laser-visual-inertial odometry and mapping system~\cite{zhang2018laser} adopts a sequential pipeline to estimate motion in a coarse to fine manner. A visual-inertial motion estimation system is followed by a scan matching module to further refine the pose. 

Some other works propose tightly-coupled sensor fusion for lidar odometry. LIO-SAM~\cite{shan2020lio} solves a factor graph including lidar odometry factor, IMU preintegration fator, GPS factor and loop closure factor. The lidar odometry factor is formed by the relative pose constraint which is computed from scan-matching. Since there is no scan-to-map registration, the drifting error will be accumulated fast and the mapping result relies on the loop closure factor and the GPS factor in outdoor. LIO-mapping~\cite{ye2019tightly} is a tightly-coupled lidar inertial odometry and mapping framework. The measurements from lidar and IMU are jointly optimized to achieve comparable or higher accuracy than LOAM system. However, the tightly-coupled lidar odometry methods requires suffcient motion excitation to IMU for bias estimation and the optimization framework is computationally heavy. 

Considering the balance between efficiency and accuracy, we adopt the idea of loosely-coupled approach as in LOAM for lidar odometry and mapping. We then apply a few effective improvements to enhance the robustness and accuracy. In addition, those works above mainly focus on pose estimation. Since the map accuracy is crucial for occlusion detection and realistic rendering in AR systems, we design a global optimization and stable mapping module to ensure both the global and local consistency of AR Maps.

\section{Backpack scanning system}
\subsection{Hardware Design}
\begin{figure}
\begin{subfigure}{0.9\textwidth}
  \includegraphics[width=\textwidth]{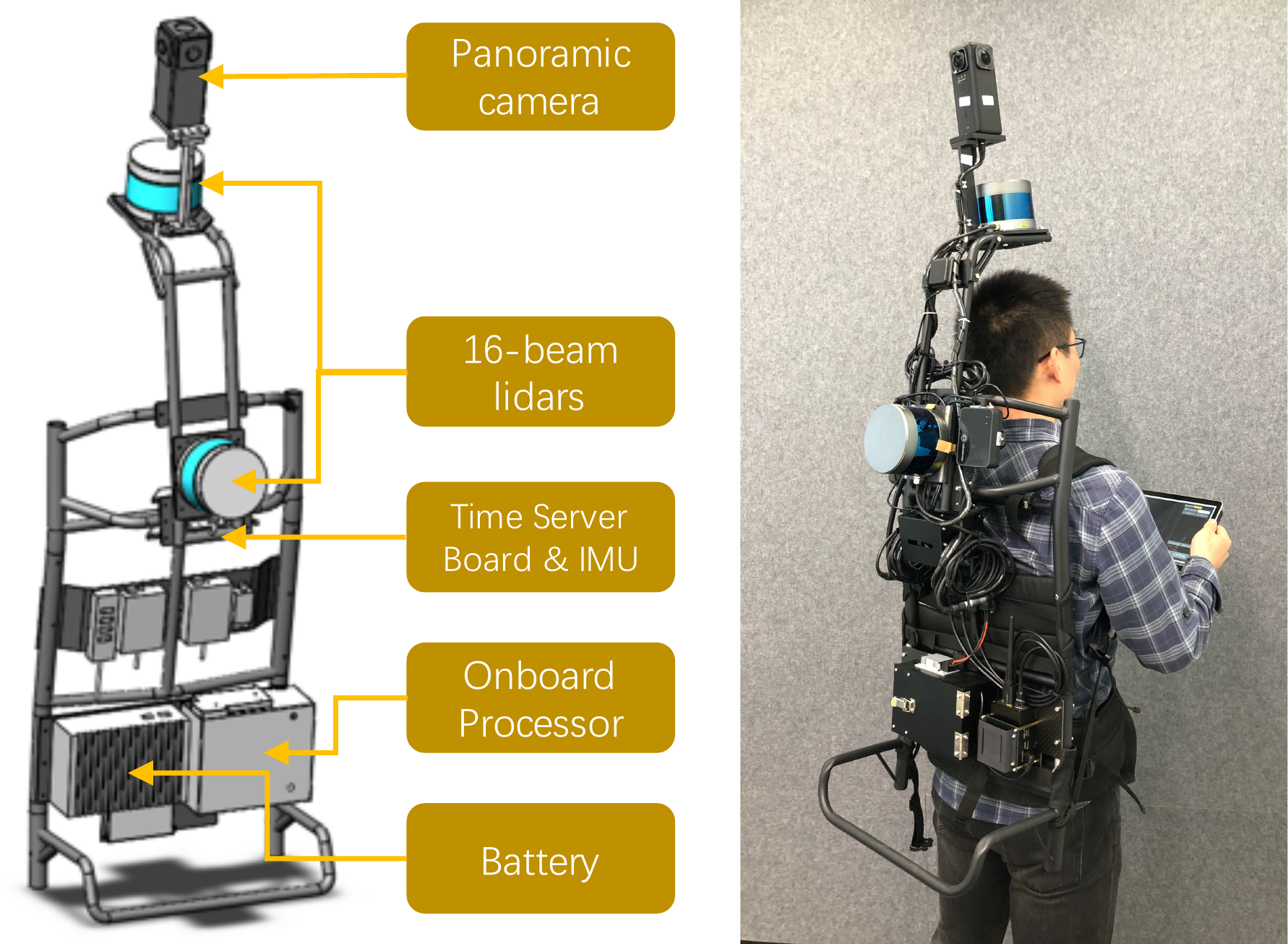}\hfill
\end{subfigure}\hfil
\caption{The hardware design and components of our backpack system (left) and an operator carrying the backpack and a touch pad for data capture (right).}\label{fig:backpack_system}
\end{figure}

Our backpack scanning system is designed for AR mapping. As shown in Fig.~\ref{fig:backpack_system}, it consists of two 16-beams RoboSense lidars, an MTi-3 AHRS IMU and a Teche 360 Anywhere panoramic camera with 4 fisheye lens. All the sensors are synchronized by a time server. The angle between the scanning planes of two lidars is about 25 degrees. As the lidar spins continuously, it rotates 360 degrees within 0.1 seconds and generate 75 data packets. The timestamp of each packet is also synchronized with the time server. The time server sends pulse signal in 100 Hz to trigger the measurement of IMU.

While scanning the environment, the operator carries the backpack system and a touch pad for controlling. When the color images need to be captured, the operator stops walking and presses a ``capture'' button on the pad to take 4 images from lens of the panoramic camera. The data from lidars are fused with IMU measurements to generate the point cloud map and 6-DOF poses at each scan. The dense depth map for each color image will be produced from the the point cloud map, lidar poses and the extrinsic parameters between lidar and camera lens. 


\subsection{Unified and Efficient Multi-Sensor Calibration}
\begin{figure}[t!]
\begin{subfigure}{0.3\textwidth}
  \includegraphics[width=\textwidth]{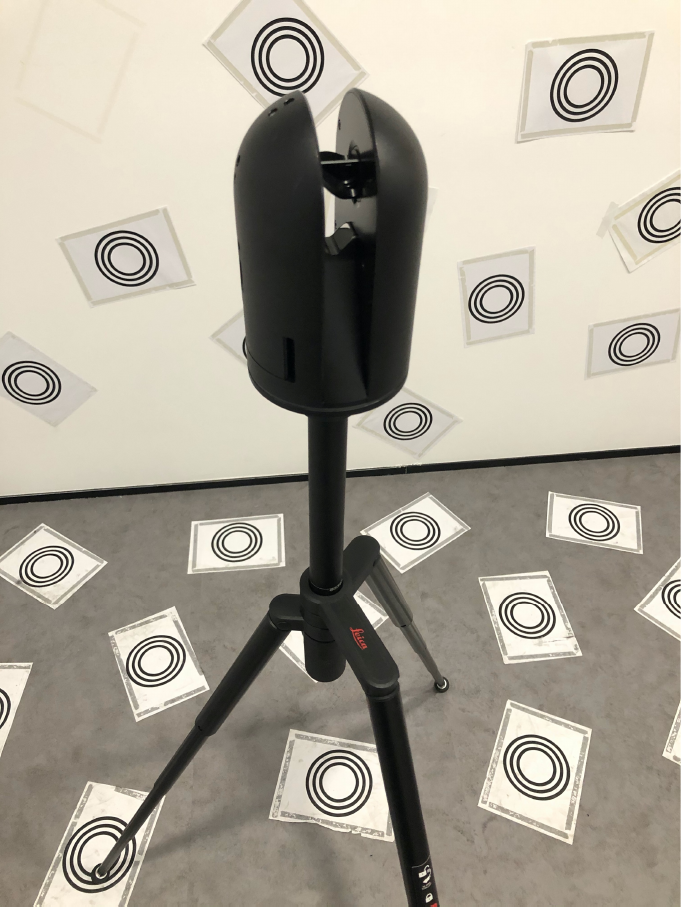}\hfill
\caption{}
\end{subfigure}\hfil
\begin{subfigure}{0.66686\textwidth}
\includegraphics[width=\textwidth]{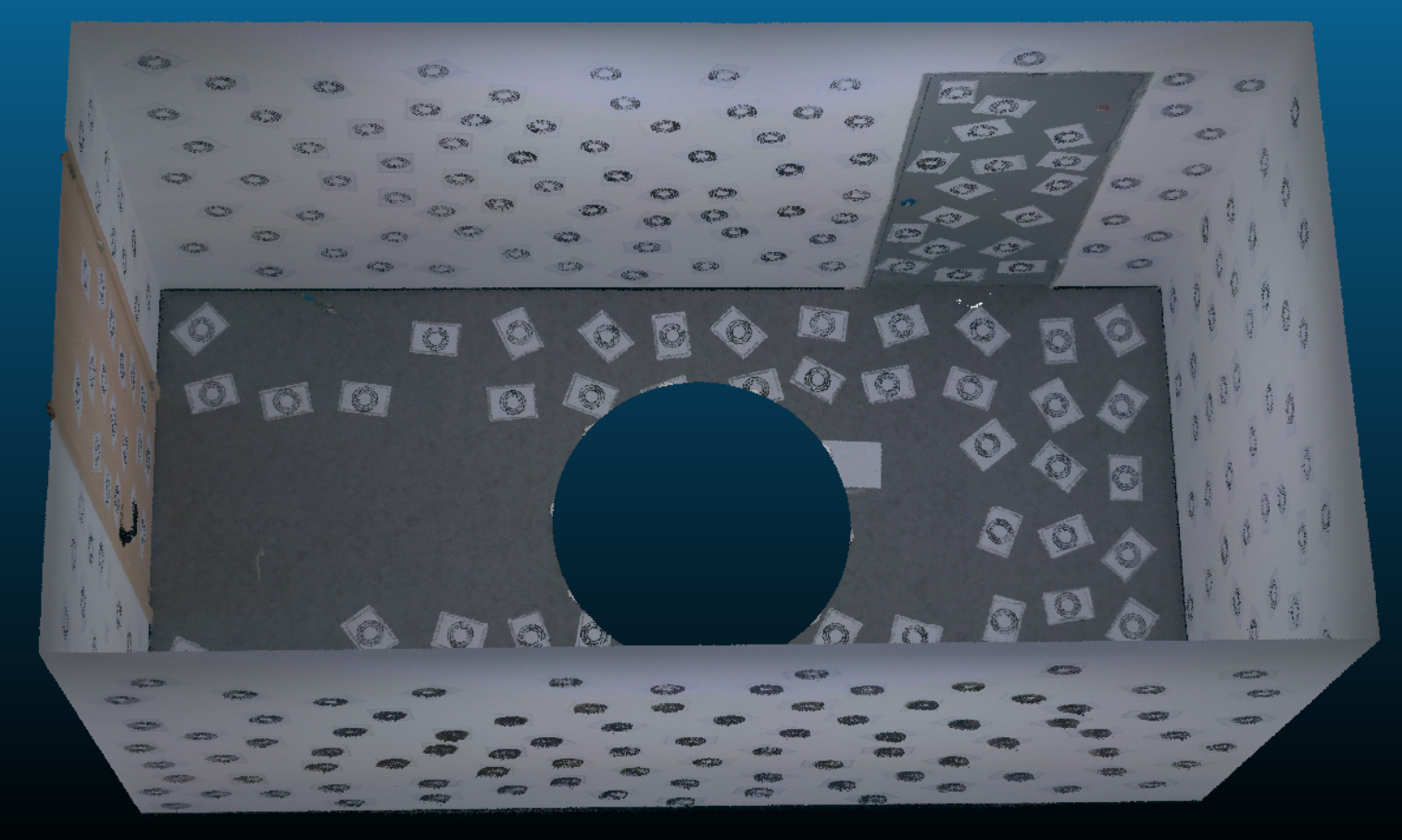}\hfill
\caption{}
\end{subfigure}\hfil
\caption{(a) The high-accuracy rotating laser scanner Leica BLK 360 mounted on a tripod. (b) The colored dense point cloud generated by capturing at one spot with Leica BLK 360. Notice that the empty area on the ground is caused by the occlusion of the tripod.}
\label{fig:leica_calibration_room}
\end{figure}

We use the method proposed by Furgale et al,~\cite{furgale2013unified} to calibrate the intrinsic parameters of the panoramic camera and the IMU. Then we present an approach to efficient extrinsic calibration of our backpack scanning system. Firstly, as shown in Fig~\ref{fig:leica_calibration_room}, we setup a calibration environment (a 4.8 m $\times$ 3.4 m $\times$ 3 m room in our case) by placing CCTag markers~\cite{calvet2016detection} densely on the wall. There is no strict requirement on the markers' positions but the distance between neighboring markers is generally less than 20 cm. We then use a high-accuracy laser scanner Leica BLK360~\cite{leica360} to reconstruct the colored dense point cloud $D$ of the calibration room. All marker positions can be determined by detecting CCtags in rendered images of the room or manual labeling. In this way, we have a sparse reconstruction $S$ of the room including all markers' 3D positions. Note that the calibration environment needs to be set up only once. 

Using the calibration room as a common reference $R$, we estimate the transformation from each lidar and camera lens to $R$ and then the relative poses between different sensors can be easily obtained. Firstly, we use 3D Delaunay triangulation to connect neighboring 3D markers in $S$. For a specific camera $C_i$, we detect the CCTags' positions in its captured image $I_i$ and apply 2D Delaunay triangulation~\cite{lee1980two} to connect nearby markers in $I_i$. Then we select a 2D triangle $\delta$ and greedily search for its corresponding 3D triangle $\Delta$ in $S$ by solving a Perspective-n-Point problem with RANSAC algorithm~\cite{li2012robust}. After an optimal 2D-3D triangle match is determined, the transformation $T^{C_i}_R$ from the camera frame to $R$ can be estimated using all inliers from the RANSAC process. Finally, for a specific lidar $L_i$, we estimate the transformation $T^{L_i}_R$ from lidar frame to $R$ by using Generalized-ICP~\cite{segal2009generalized} to register $L_i$'s scan and dense point cloud $D$. The initial pose $\hat{T}^{L_i}_R$ can be obtained by multiplying $T^{C_i}_R$ with the manually measured transformation $\hat{T}^{L_i}_{C_i}$ from $L_i$ to $C_i$.

\section{AR Mapping System}
\subsection{System Overview}
The raw data generated from our backpack scanning device includes the point cloud per scan from lidars, IMU measurements and images from 4 fisheye lens of the panoramic camera. The raw data is then fed into an AR mapping system to generate AR Maps. Firstly, a lidar odometry system produces unskewed point cloud and 6-DOF pose for each scan. In order to correct the drifting error accumulated during the odometry, the results of the lidar odometry are then optimized in a global optimization module. This module is designed to ensure both global and local consistency of point cloud map. After that, a stable lidar mapping is carried out to construct the final point map with an map fusion strategy for outlier filtering. Finally, the pose of color images is interpolated from lidar poses and the dense depth maps are rendered from the point cloud map. We then elaborate the individual modules in the following. The pipeline of our AR Mapping system is illustrated in Fig.~\ref{fig:armapping_pipeline}.

\subsection{Lidar Odometry System with Improved Feature Selection}

To be self-contained, we give a brief review on the LOAM system~\cite{zhang2014loam} and then propose our improvements. For each scan, the edge and planar features are extracted based on a ``curvature'' score computed from the distance between neighboring points. Then scan-to-scan motion is estimated by solving a point-to-line or point-to-plane iterative closest point (ICP) problem. A map consisting of features is built online and each scan is further registered to the feature map for pose refinement. The measurements from an IMU are used to make motion prediction and remove the scan's distortion introduced by lidar's movement. 

\begin{figure}[ht]
\includegraphics[width=0.5\textwidth]{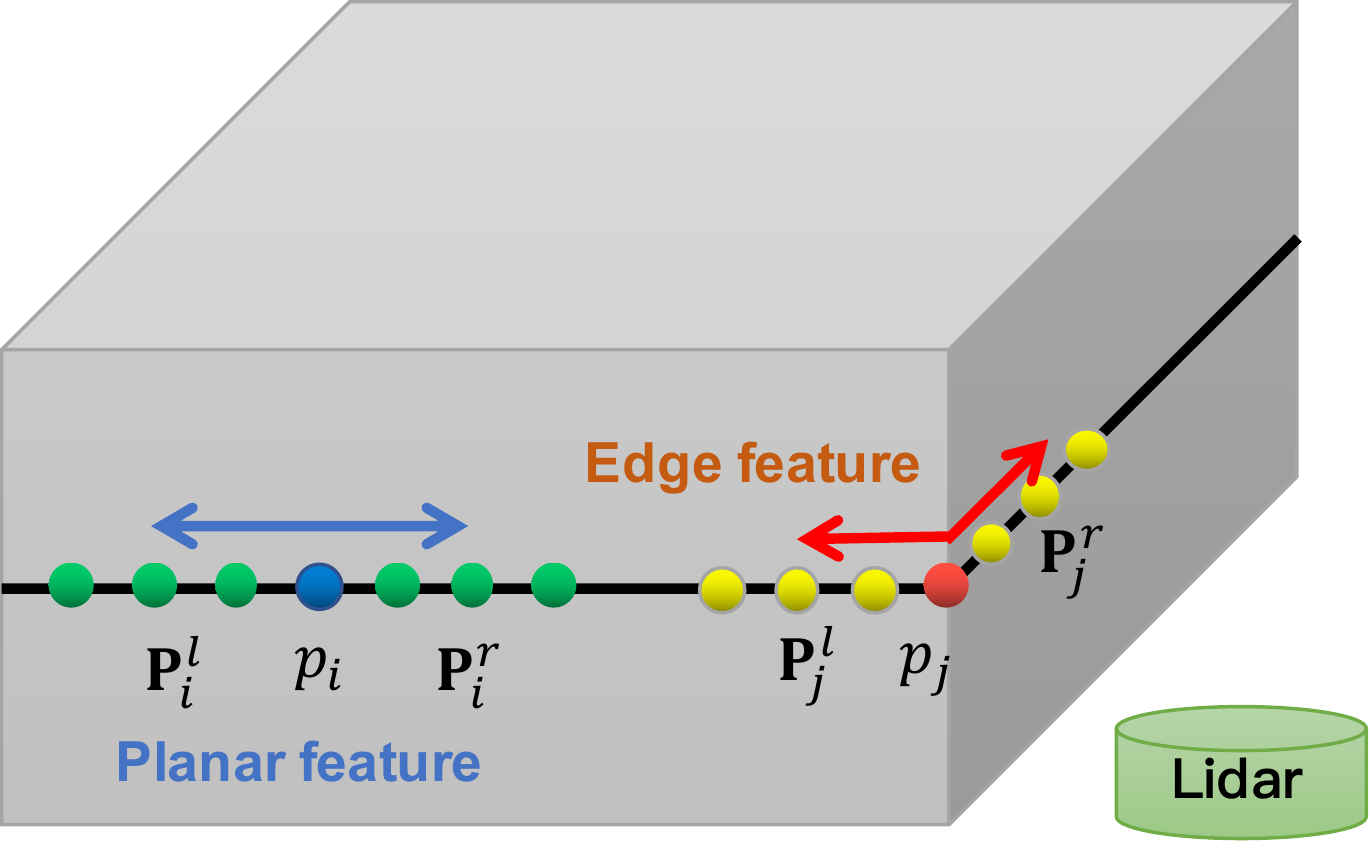}
\caption{Feature selection based on the curvature scores and curve angle.}
\label{feature1}
\end{figure}

For the unskewed point cloud of each scan, we extract the edge and planar features. As described in ZhangJi's work~\cite{zhang2014loam}, the features are selected based on a ``curvature'' score which measures the smoothness of local surface as following,
\begin{equation}
\label{equ:curvature_score}
    c_1=\frac{1}{|P_n|\cdot \left\|\mathbf{p}_i\right\|} \left\|{\sum_{j\in S}(\mathbf{p}_j - \mathbf{p}_i)}\right\|,
\end{equation}
where $\mathbf{p}_i$ is $i_{th}$ point in the scan, $\mathbf{p}_j$ belongs to a point set $P_n$ which consists of the neighboring points on the same scan line of $\mathbf{p}_i$. $c_1$ is the curvature score. As shown in Equation~\ref{equ:curvature_score}, the surface smoothness is measured by the sum of distance from one point's neighbors to itself. The intuition behind is that the distance summation will be small if the neighbors lie on the same plane with $\mathbf{p}_i$. Otherwise, the score is large if $\mathbf{p}_i$ is on an edge, However, if there is occlusion between neighboring points or the points' distribution is noisy,  the curvature score will fail to convey the surface smoothness information.
To address this problem, we add another score to improve the curvature model as following,
\begin{equation}
\label{equ:curve_angle}
    c_2 = {\langle\frac{1}{|P_l|}\sum_{\mathbf{p}_l\in P_l}(\mathbf{p}_l - \mathbf{p}_i), \frac{1}{|P_r|}\sum_{\mathbf{p}_r\in P_r}(\mathbf{p}_r - \mathbf{p}_i)\rangle},
\end{equation}
where $P_l$ and $P_r$ are the point sets on the left and right to $\mathbf{p}_i$ in the same scan line respectively. $\langle \mathbf{v}_1, \mathbf{v}_2 \rangle$ represents the angle between the vector $\mathbf{v}_1$ and $\mathbf{v}_2$. Intuitively, $c_2$ represents the curve angle of the scan line at $\mathbf{p}_i$. As shown in Fig~\ref{feature1}, $c_2$ will be large for planar features and small for the edge features. Therefore, a point $p_i$ can be classified to edge or planar features based on the criteria as follows,
\begin{equation}
    p_i\in
    \begin{cases}
      F_e, & \text{if}\ c_1 > {\alpha}_e, c_2 < {\beta}_e\\
      F_p, & \text{if}\ c_1 < {\alpha}_p, c_2 > {\beta}_p\\
    \end{cases}
\end{equation}
where $F_e$ and $F_p$ are the sets of edge and planar features respectively. ${\alpha}_e(=0.005)$, ${\beta}_e(=2.09 rad)$, ${\alpha}_p(=0.0002) $ and ${\beta}_p(=2.62 rad)$ are the thresholds. 

\begin{figure}[t]
\begin{subfigure}{0.3\textwidth}
\includegraphics[height=2cm]{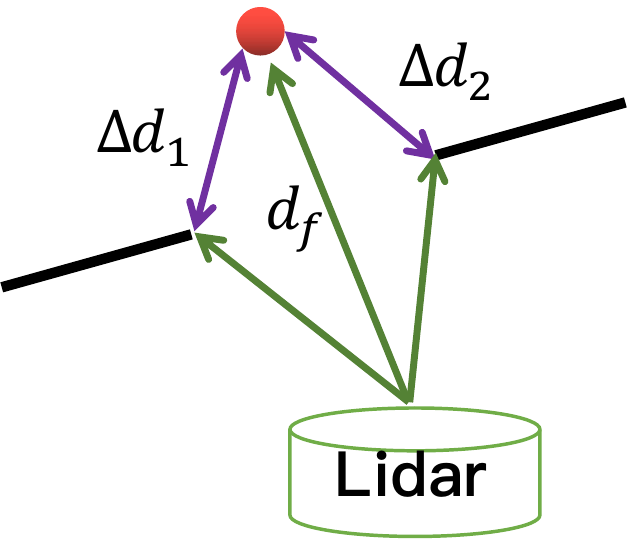}
\caption{}\label{feature_filter1}
\end{subfigure}
\begin{subfigure}{0.3\textwidth}
\includegraphics[height=2cm]{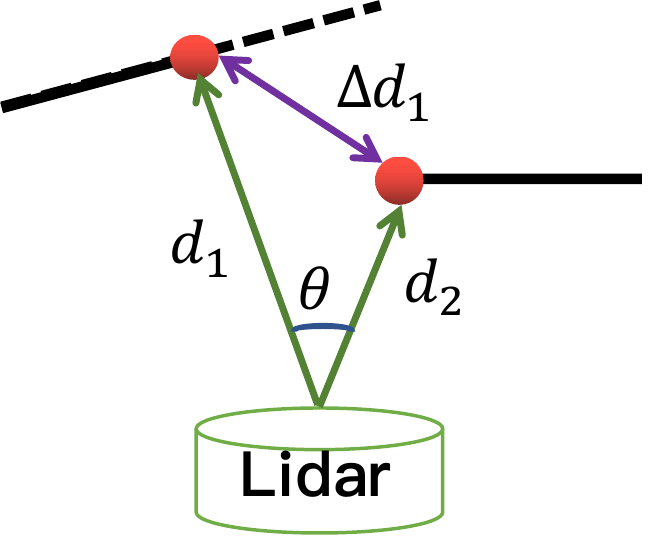}
\caption{}\label{feature_filter2}
\end{subfigure}
\begin{subfigure}{0.3\textwidth}
\includegraphics[height=2cm]{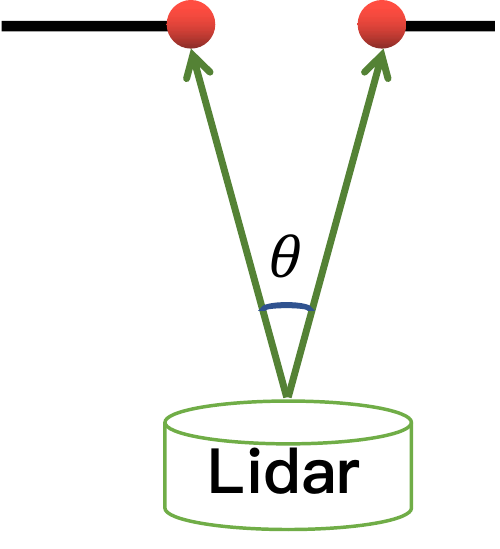}
\caption{}\label{feature_filter3}
\end{subfigure}
\caption{Outlier features should be removed in cases: (a) outlier point, (b) depth discontinuity, (c) occluded regions.} 
\label{fig:feature_filter}
\end{figure}

\begin{figure}[!t]
\begin{subfigure}{0.48\textwidth}
\includegraphics[width=\textwidth]{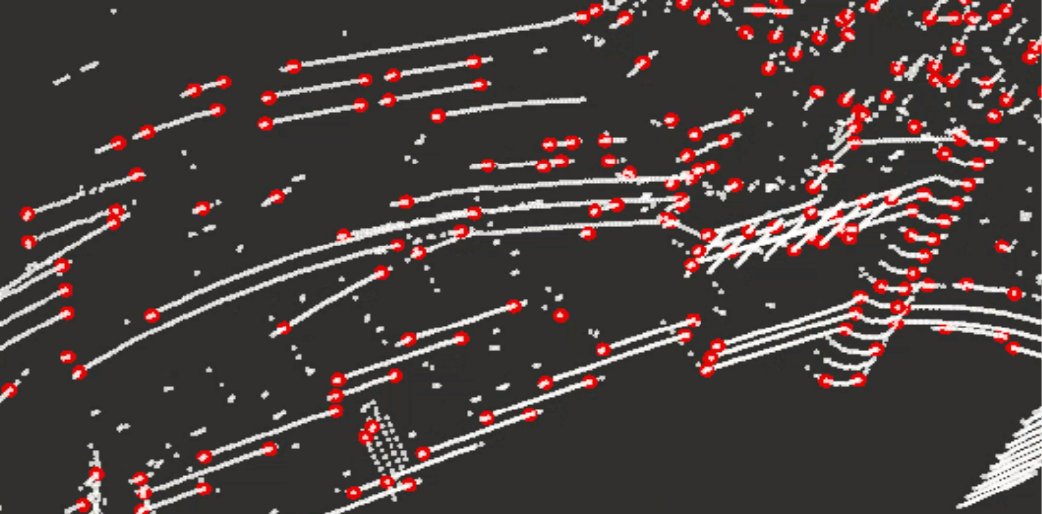}
\caption{Without feature filter}\label{feautre_filter_without}
\end{subfigure}
\begin{subfigure}{0.48\textwidth}
\includegraphics[width=\textwidth]{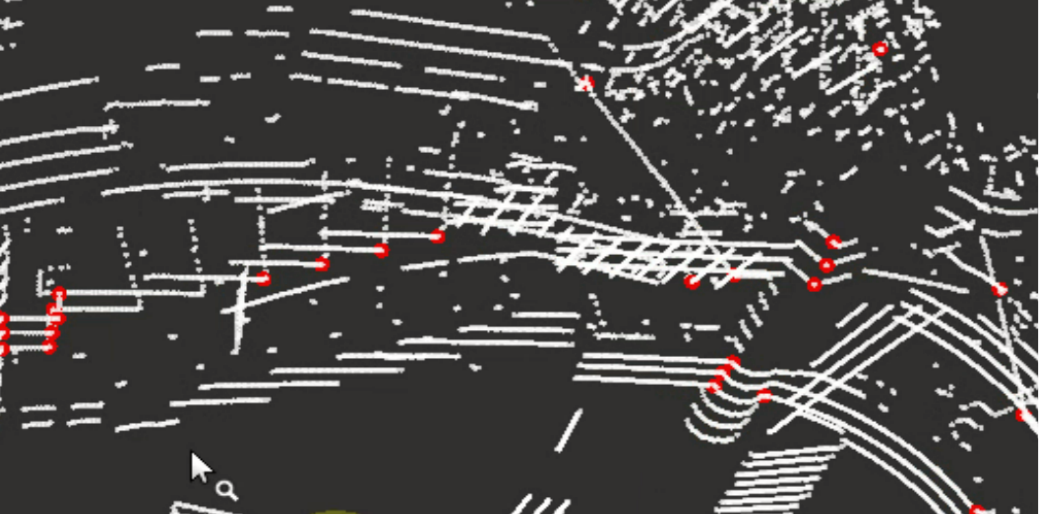}
\caption{With feature filter}\label{feature_filter_with}
\end{subfigure}
\caption{A demonstration of edge features (coloered in red) in one scan with and without feature filter. It can be seen that a large number of false positive edge features are removed and only those points at real edges are kept.}\label{fig:feature_filter_compare}
\end{figure}

However, the curvature score and curve angle may both fail to describe a reliable feature in the following cases: 1) outlier point(Fig.~\ref{feature_filter1}); 2) depth discontinuity(Fig.~\ref{feature_filter2}); 3) occluded regions(Fig.~\ref{feature_filter3}). As shown in Fig.~\ref{fig:feature_filter}, the point in red will be falsely classified as edge feature based on $c_1$ and $c_2$. Therefore, we apply feature filtering strategy for each case and remove all outlier points in above cases: 
\begin{equation}
    p_i\in
    \begin{cases}
      \text{outlier points}, & \text{if}\ \Delta d_1 / d_f > r_1 , \Delta d_2 / d_f > r_1\\
      \text{depth discontinuity}, & \text{if}\ |d_1 - d_2| > {\gamma}_1, \theta < {\theta}_1\\
      \text{occluded regions}, & \text{if}\ \theta > {\theta}_2\\
    \end{cases}
\end{equation}
where $\Delta d$ is the distance between neighboring points on the same scan line. $d_f$ is the distance from lidar to the scan points. $\theta$ is the angle between vectors pointing from lidar to neighboring points. $r_1(=0.014)$, ${\gamma}_1(=0.33)$, ${\theta}_1(=5.72 rad)$ and ${\theta}_2(=0.57 rad)$ are the thresholds. As shown in Fig.~\ref{fig:feature_filter_compare}, a large number of false positive edge features are removed by the feature filtering.

\subsection{Global Optimization}
\begin{figure}[!t]
\begin{subfigure}{0.48\textwidth}
\includegraphics[width=\textwidth]{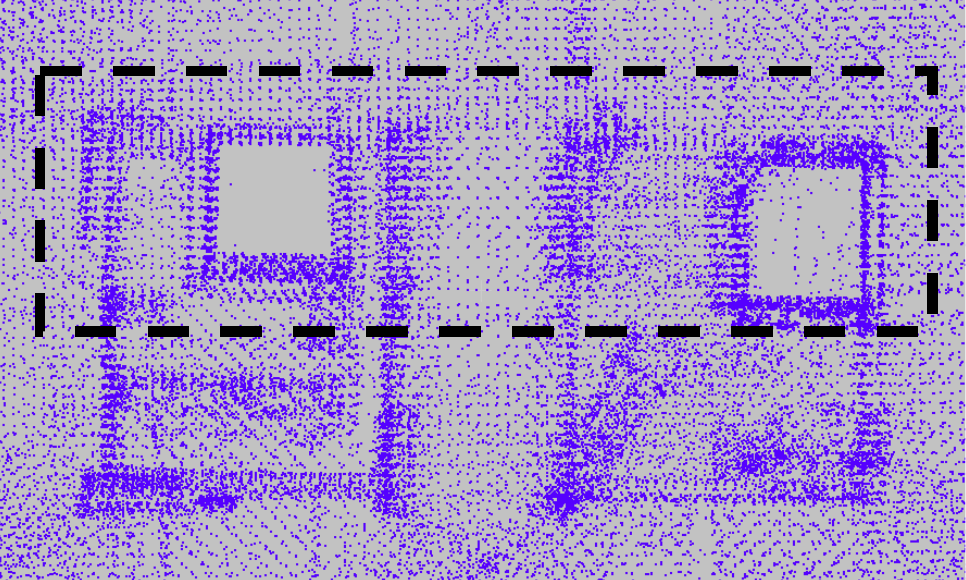}
\caption{Pose Graph only}\label{global_opti_pose_only}
\end{subfigure}
\begin{subfigure}{0.48\textwidth}
\includegraphics[width=\textwidth]{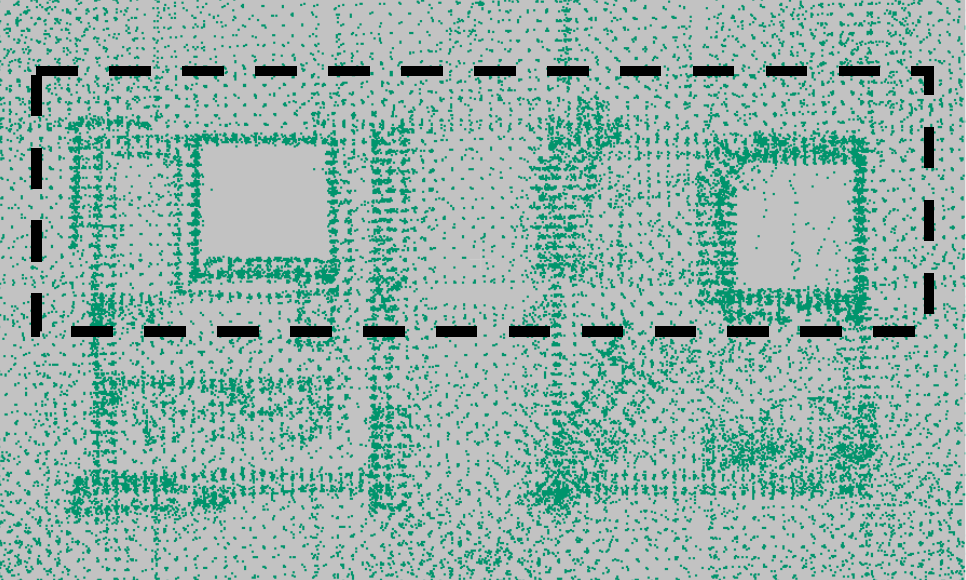}
\caption{Global Optimization}\label{global_opti_ba}
\end{subfigure}
\caption{Top view of a detail in point cloud map. It shows the comparison between with (left) and without (right) map consistency constraints in global optimization. Notice the blurry edges in the left and the sharp structure in the right.}\label{fig:global_opti_compare}
\end{figure}
The point cloud map generated from lidar odometry usually suffers from accumulated drifting error especially in large scale scenes. This results in inconsistent map at loop points where the backpack scans twice. We adopt a submap-based loop closing and global optimization strategy to refine the trajectory while enforcing the map consistency. Traditional lidar odometry system~\cite{shan2020lio,shan2018lego} utilizes a pose graph to optimize the lidar's trajectory. However, in this way it only enforces the pose consistency at loop points while ignores the map consistency globally. To resolve this issue, we follow an approach similar to sparse surface adjustment~\cite{ruhnke2012highly} and optimize the final trajectory with constraints of poses and map points.

After processing lidar odometry, we save the undistorted point cloud and estimated pose for each scan. Assuming the drifting error is small in a short period of time, we construct submaps consisting of a small number of consecutive scans. The pose of the submap is the scan pose in the middle of the trajectory. For a specific submap $s_i$, we search for the spatially nearby submaps $S_n$. We then remove the ones close to $s_i$ temporally in $S_n$. Finally, we get a set $S_c$ consisting of the submap candidates for loop closing. For each candidate $s_c$, we use generalized-ICP~\cite{segal2009generalized} to register $s_c$ and $s_i$ and obtain the relative pose $T^{s_c}_{s_i}$. For all point correspondences within 0.5 m, we compute the fitness score which is the sum of squared distance from source to target. The registered submaps with fitness score larger than 0.1 will be rejected. A pose graph can be constructed using the pose constraints between the consecutive submaps and submap pairs from the loop detection. In this way, we construct a pose graph consisting of relative pose constraints as follows,
\begin{equation}
\label{equ:cost_pose}
\begin{aligned}
    E_{pose} =& \sum_{T_i, T_{i+1} \in S_p}{T^{-1}_{i+1}\cdot T_{i} \cdot \delta T^{odom}_{i,i+1}} \\
               &+\sum_{(T_j,T_k) \in S_l}{T^{-1}_k \cdot T_j \cdot \delta T^{loop}_{j,k}}
\end{aligned}
\end{equation}
where $T_i$ stands for the transformation from world frame to submap $i$'s frame. $S_p$ is the set of all submaps and $S_l$ is the set of all submap pairs generated from loop detection. $\delta T^{odom}_{i,i+1}$ is the relative pose between consecutive submaps from lidar odometry and $\delta T^{loop}_{j,k}$ is the relative pose between submap pairs computed from loop closing.

To obtain a consistent point cloud model, Ruhnke et al,~\cite{ruhnke2012highly} propose to do a joint optimization over sensor pose and the positions of surface points, which increases variables significantly and is computationally heavy. Inspired by this idea, we then add the point-to-plane constraints between submaps to enforce the map consistency. For each point $\mathbf{p}_i$ in submap $s_i$, the closest 5 neighboring points are found using a k-d tree and used to fit plane parameters $(\mathbf{n}, d)$. If the distance from any point to the plane is larger than 0.1 m, we reject the fitted plane. Finally, we use all the constraints between point to valid planes. The cost function is presented below,
\begin{equation}
\label{equ:cost_points}
    E_{loop} = \sum_{(T_j,T_k) \in S_l}{\mathbf{n}_j \cdot (T^{-1}_j \cdot T_k \cdot \mathbf{p}_k) + d_j},
\end{equation}
where $\mathbf{p}_k$ is a point in submap $k$, $(\mathbf{n}_j, d_j)$ is the corresponding plane parameters fitted in submap $j$. We then optimize the submap poses by minimizing the total cost function,
\begin{equation}
\label{equ:cost_total}
    E_{total} = \argmin_{T_i \in S_p} (w_1 E_{pose} + E_{loop}),
\end{equation}
where $w_1$ is a balancing weight. Finally, all the poses for scan will be adjusted according to the optimized submap poses. Fig~\ref{fig:global_opti_compare} shows the comparison between pose graph optimization and our global optimization. The map is more consistent after the global optimization.



\subsection{Stable Map Construction}
\begin{figure}[t!]  
\begin{subfigure}{0.49\textwidth}
\includegraphics[width=\linewidth]{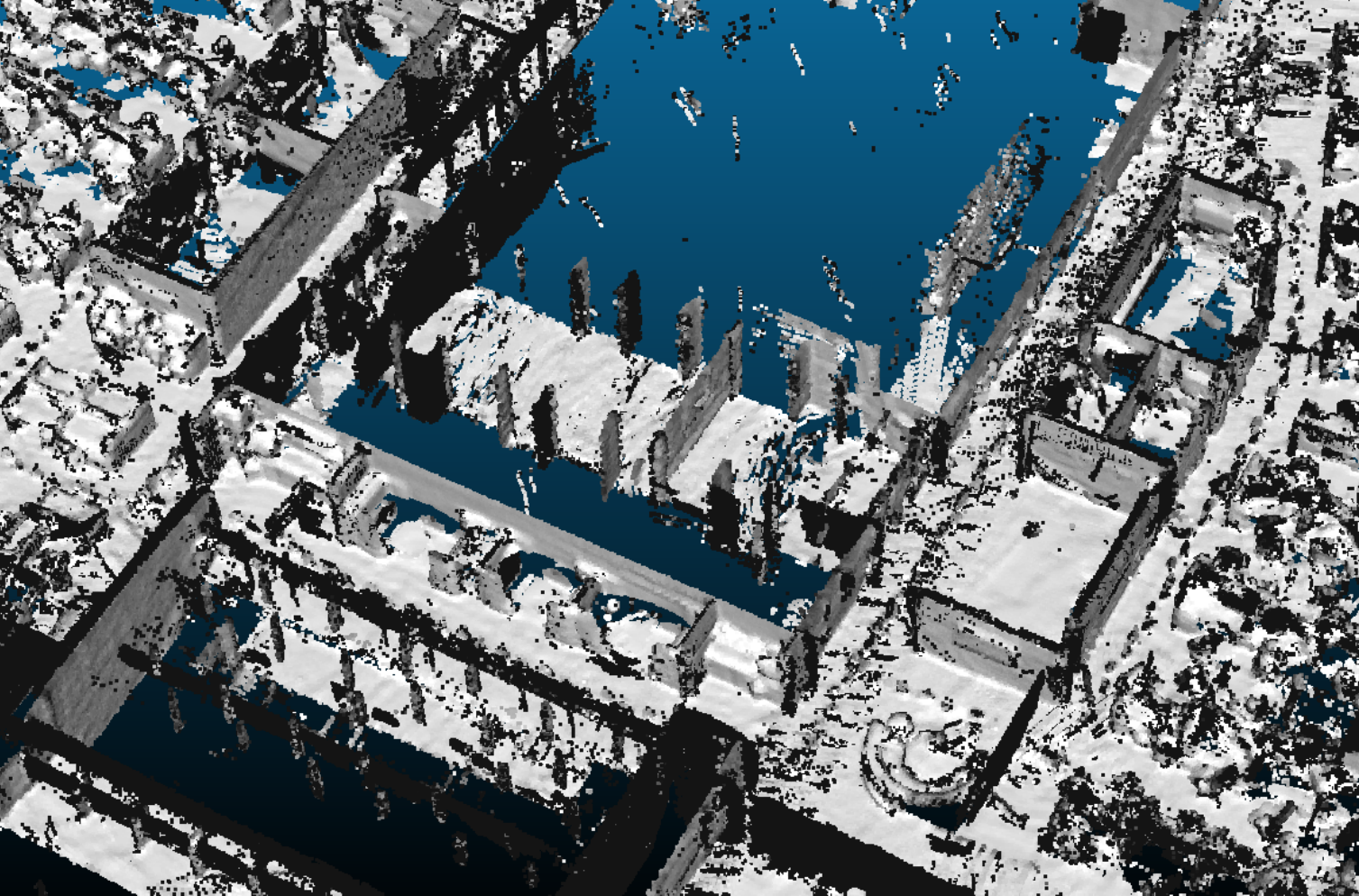}
\caption{}\label{fig:stable_map_noisy}
\end{subfigure}
\begin{subfigure}{0.49\textwidth}
\includegraphics[width=\linewidth]{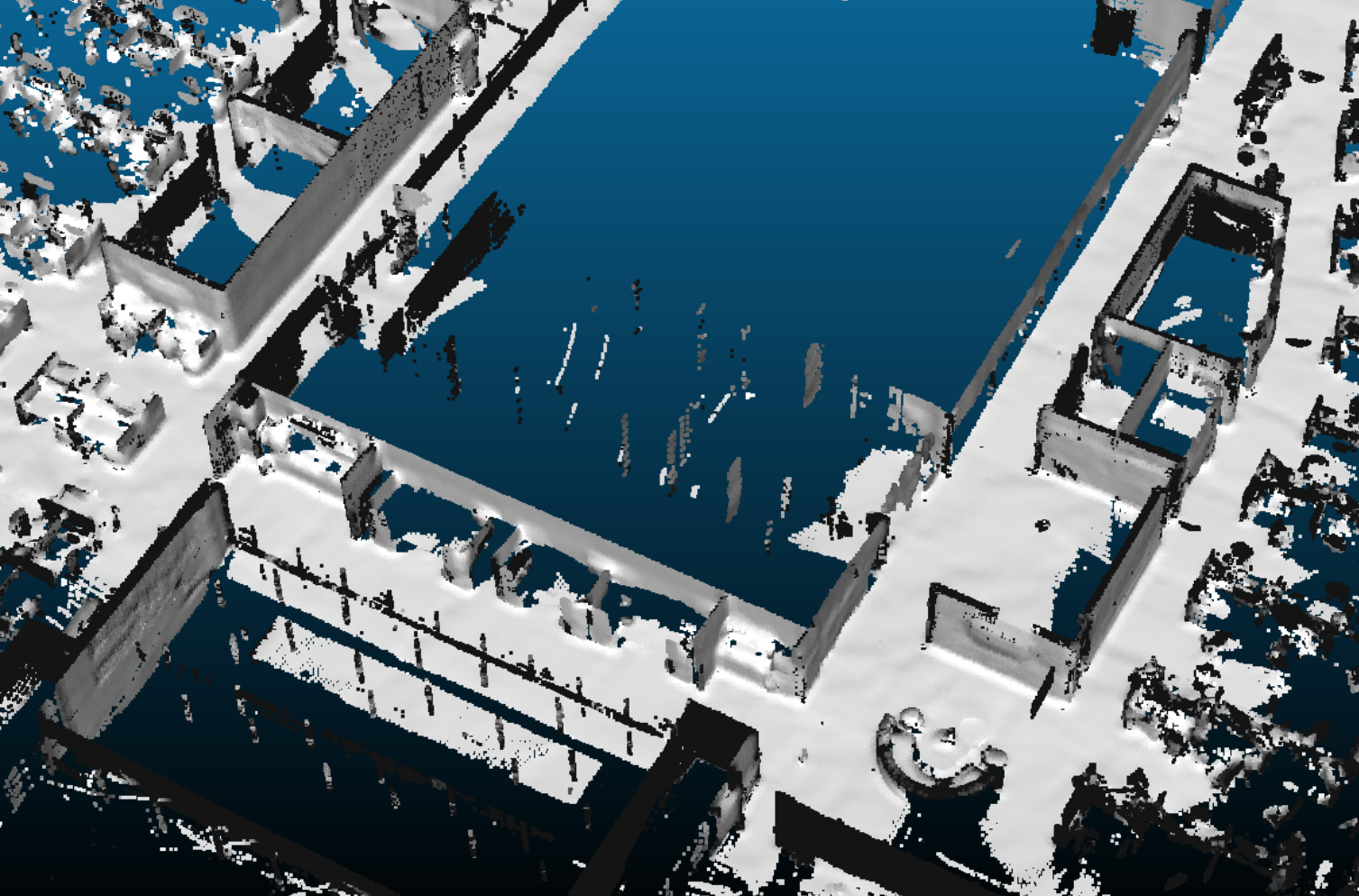}
\caption{}\label{fig:stable_map_clean}
\end{subfigure}
\caption{We show the point cloud map constructed in a busy office with people moving around often. The left shows the result of simply stitching all point clouds and the right shows the stable mapping result.} 
\label{fig:stable_mapping_compare}
\end{figure}

Once the global optimization is finished, the complete point cloud map is then constructed by stitching all scans' point clouds using their poses. In the original LOAM system, the map consisting of only sparse feature points is maintained online. The feature map is divided into cubes with size of $d_c \times d_c \times d_c$ ($d_c = 50 m$ in implementation). After a scan's points is added into the feature map, the point cloud in the corresponding cubes are downsampled by a voxel grid filter. This filter constructs a voxel grid over the input point cloud and keeps the centroid of points in each voxel. This process is illustrated in Fig~\ref{fig:stable_mapping_compare}. Specifically, for the map point $\mathbf{p}_{k}^{m}$ in voxel $V_{k}$, the update process with input scan points $\mathbf{P}_{k}^{s}$ in $V_{k}$ can be written as the following,
\begin{equation}
\label{equ:stable_map_update_loam}
    \mathbf{p}_{k}^{m} = \frac{1}{1+|\mathbf{P}_{k}^{s}|} (\mathbf{p}_{k}^{m} + \sum_{\mathbf{p}^{s} \in \mathbf{P}_{k}^{s}}{\mathbf{p}^{s}}),
\end{equation}
Intuitively, as can be seen from Equation~\ref{equ:stable_map_update_loam}, the updated map point $\mathbf{p}_{k}^{m}$ is mainly determined by the centroid position of the input scan points $\mathbf{P}_{k}^{s}$. However, as the drifting error in pose estimation is accumulated during the odometry, the position of scan points $\mathbf{P}_{k}^{s}$ will drift as well. If we give a higher weight to the input scan points, the map point will drift away from the orginal position. We then re-design the map fusion strategy. First, we maintain a map consisting of voxels with finer resolution (= $5cm$ in our implementation). For the map point in each voxel, we use the following update policy to prevent drifting error.  
\begin{equation}
\label{equ:stable_map_update_backpack}
    \mathbf{p}_{k}^{m} = \frac{1}{N_{k}^{m}+|\mathbf{P}_{k}^{s}|} (N_{k}^{m} \cdot \mathbf{p}_{k}^{m} + \sum_{\mathbf{p}^{s} \in \mathbf{P}_{k}^{s}}{\mathbf{p}^{s}}),
\end{equation}

\begin{equation}
\label{equ:stable_map_update_point_number}
    N_{k}^{m} = N_{k}^{m} + |\mathbf{P}_{k}^{s}|,
\end{equation}
where the $N_{k}^{m}$ is the set of historical scan points input to the voxel $k$. In addition, to filter out the points on dynamic objects and noisy points, we keep the map points in voxels with sufficient observations by checking $N_{k}^{m} > \tau$. Fig.~\ref{fig:stable_mapping_compare} shows the stable mapping module is able to produce cleaner and more accurate map in highly dynamic environments.

\subsection{Image Pose Interpolation and Depth Map Rendering}
So far we have optimized poses for each scan and a complete point cloud map. We then interpolate the camera pose of captured color images according to the timestamps. A 3D model is generated from the point cloud map using Poisson surface reconstruction~\cite{kazhdan2006poisson} and used to render dense depth maps.

\section{Experiments}


\subsection{Evaluation on Point Cloud Map Accuracy}
\begin{figure}[t!]
    \centering 
\begin{subfigure}{0.48\textwidth}
  \includegraphics[width=\textwidth]{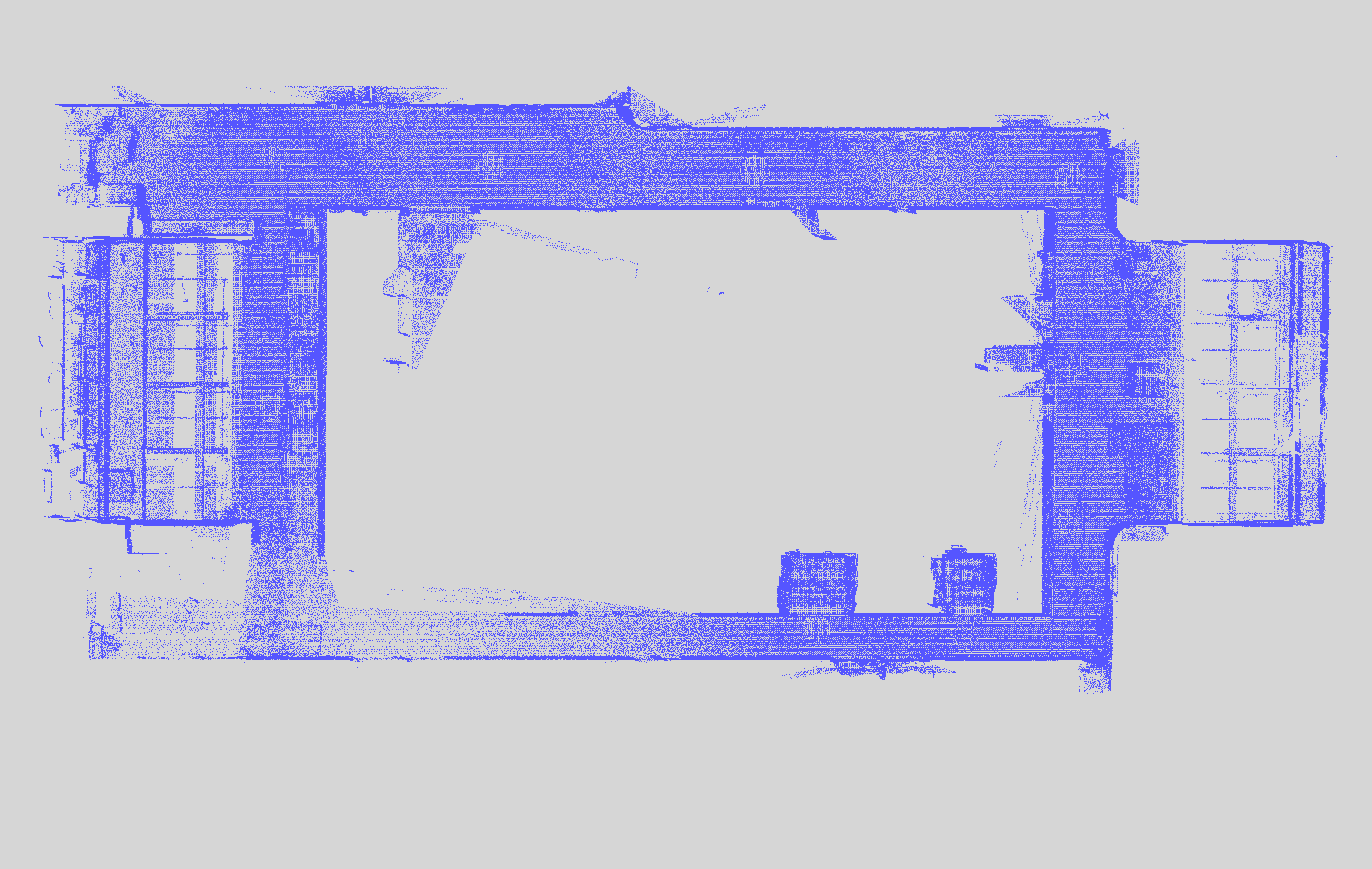}\hfill
  \label{fig:compare_leica_office}
\end{subfigure}\hfil
\begin{subfigure}{0.48\textwidth}
\includegraphics[width=\textwidth]{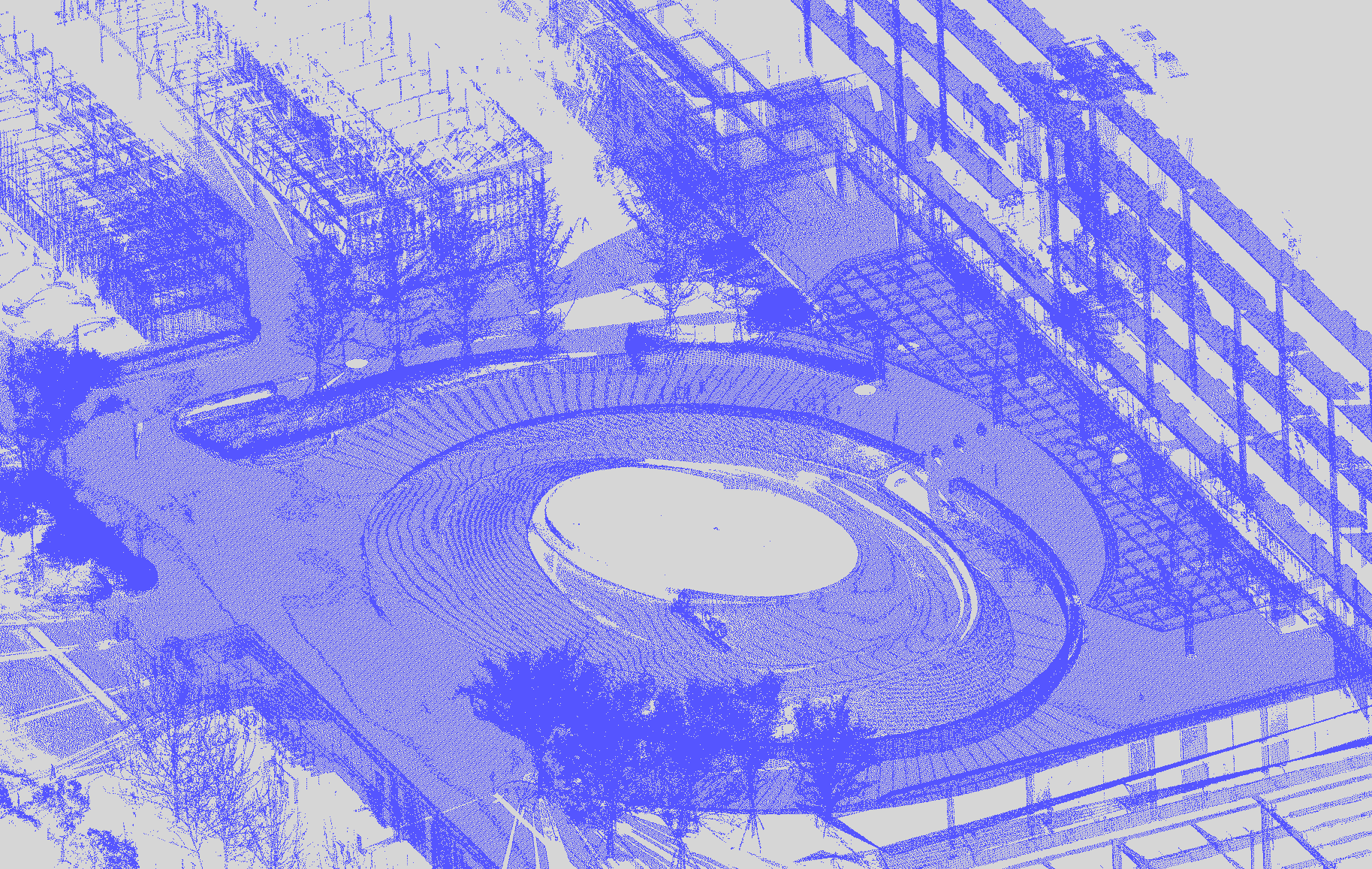}\hfill
 \label{fig:compare_leica_building}
\end{subfigure}\hfil

\begin{subfigure}{0.48\textwidth}
  \includegraphics[width=\textwidth]{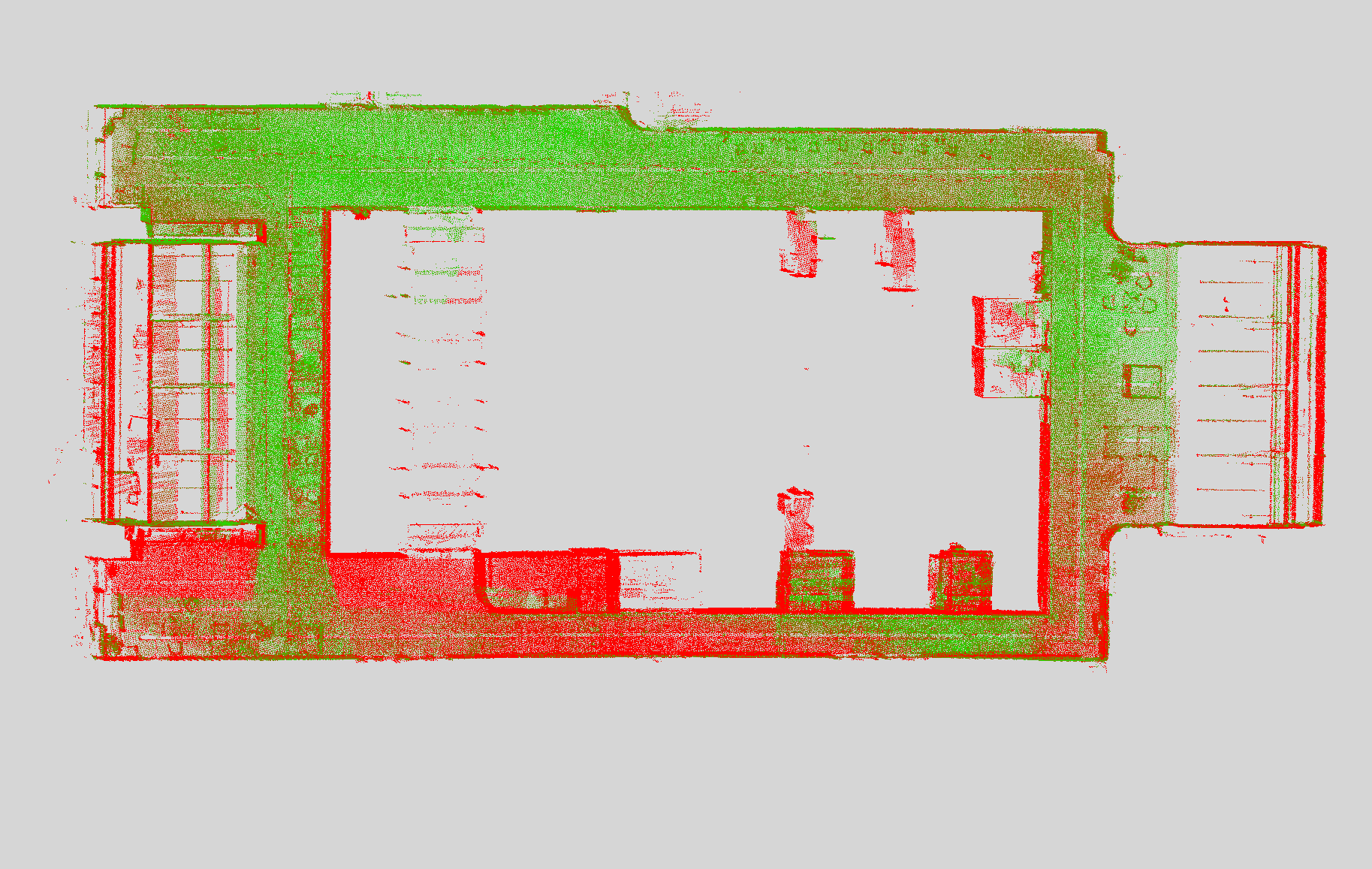}\hfill
  \caption{Office}\label{fig:compare_leica_office}
\end{subfigure}\hfil
\begin{subfigure}{0.48\textwidth}
\includegraphics[width=\textwidth]{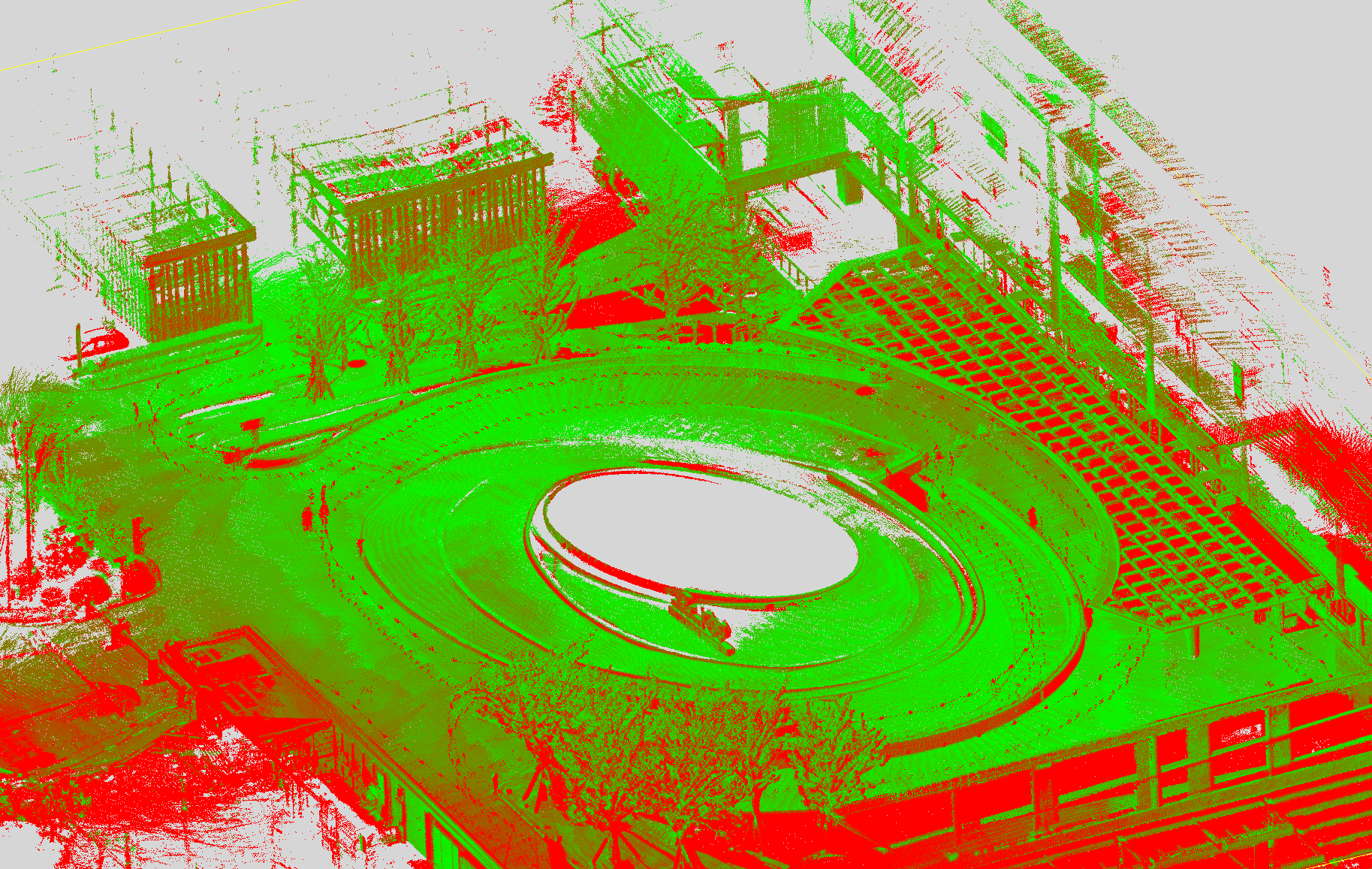}\hfill
 \caption{Campus building}\label{fig:compare_leica_building}
\end{subfigure}\hfil
\caption{The comparison of point map generated from Leica BLK360 (the first row) and our AR mapping system. (the second row).}\label{fig:compare_leica}
\end{figure}

To evaluate the mapping accuracy of our backpack system, we use a high-accuracy laser scanner Leica BLK360 to capture the point maps in different environments as ground truth. We compare the mapping results from our backpack with the ground truth maps in 3 different scenes: (a) a $45 m \times 23 m$ office floor; (b) a $100 m \times 50 m$ campus building floor; (c) a $100 m \times 60 m$ outdoor open area. Fig~\ref{fig:compare_leica} shows the point cloud maps from Leica BLK360 and AR mapping. We align the two maps by selecting corresponding points manually and conducting ICP registration. Table~\ref{tab:lidar_map_accuracy} shows the mapping accuracy is below $0.05 m$ for indoor environments and below $0.10 m$ for the outdoor environment. Table~\ref{tab:lidar_map_efficiency} shows the time of map capturing. Compared to the high-cost industrial laser scanner, the backpack scanning system produces map with comparable accuracy and much higher efficiency.

\begin{figure}
\begin{subfigure}{0.5\textwidth}
  \includegraphics[width=\textwidth]{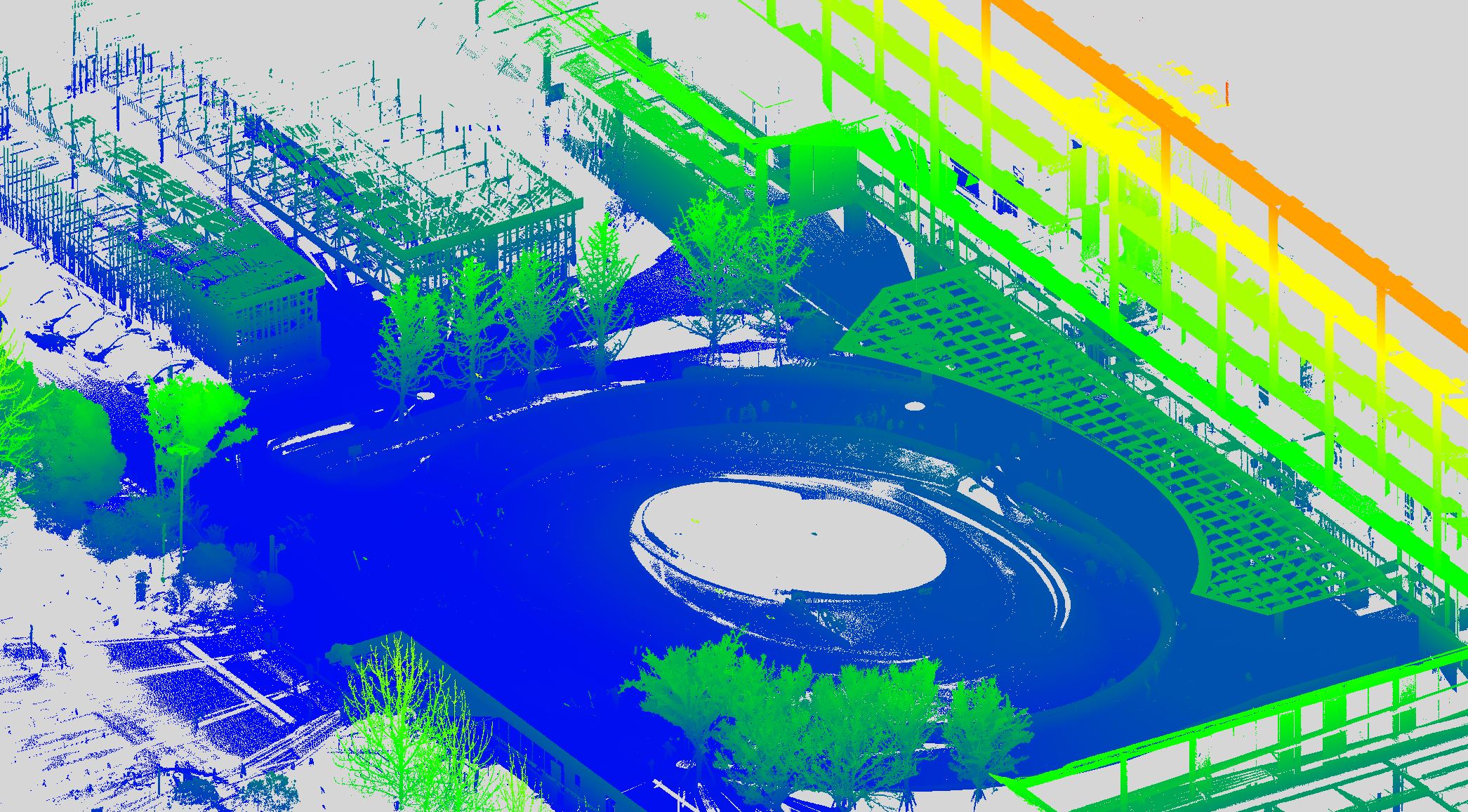}\hfill
\caption{Leica Point Cloud}
\label{fig:lio_sam_compare_leica}
\end{subfigure}\hfil
\begin{subfigure}{0.5\textwidth}
\includegraphics[width=\textwidth]{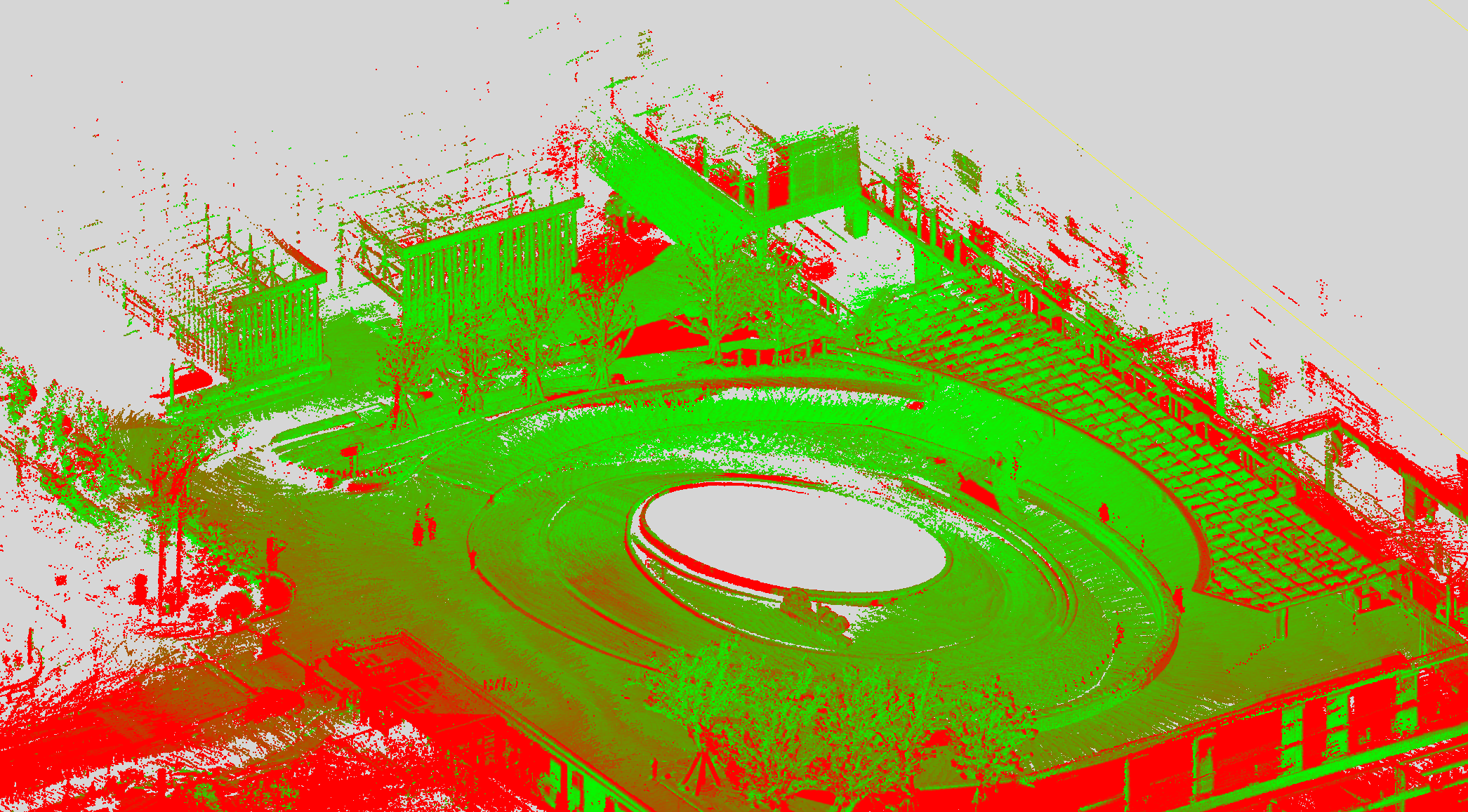}\hfill
\caption{AR Map Point Cloud}
\label{fig:lio_sam_compare_backpack}
\end{subfigure}\hfil

\begin{subfigure}{0.5\textwidth}
  \includegraphics[width=\textwidth]{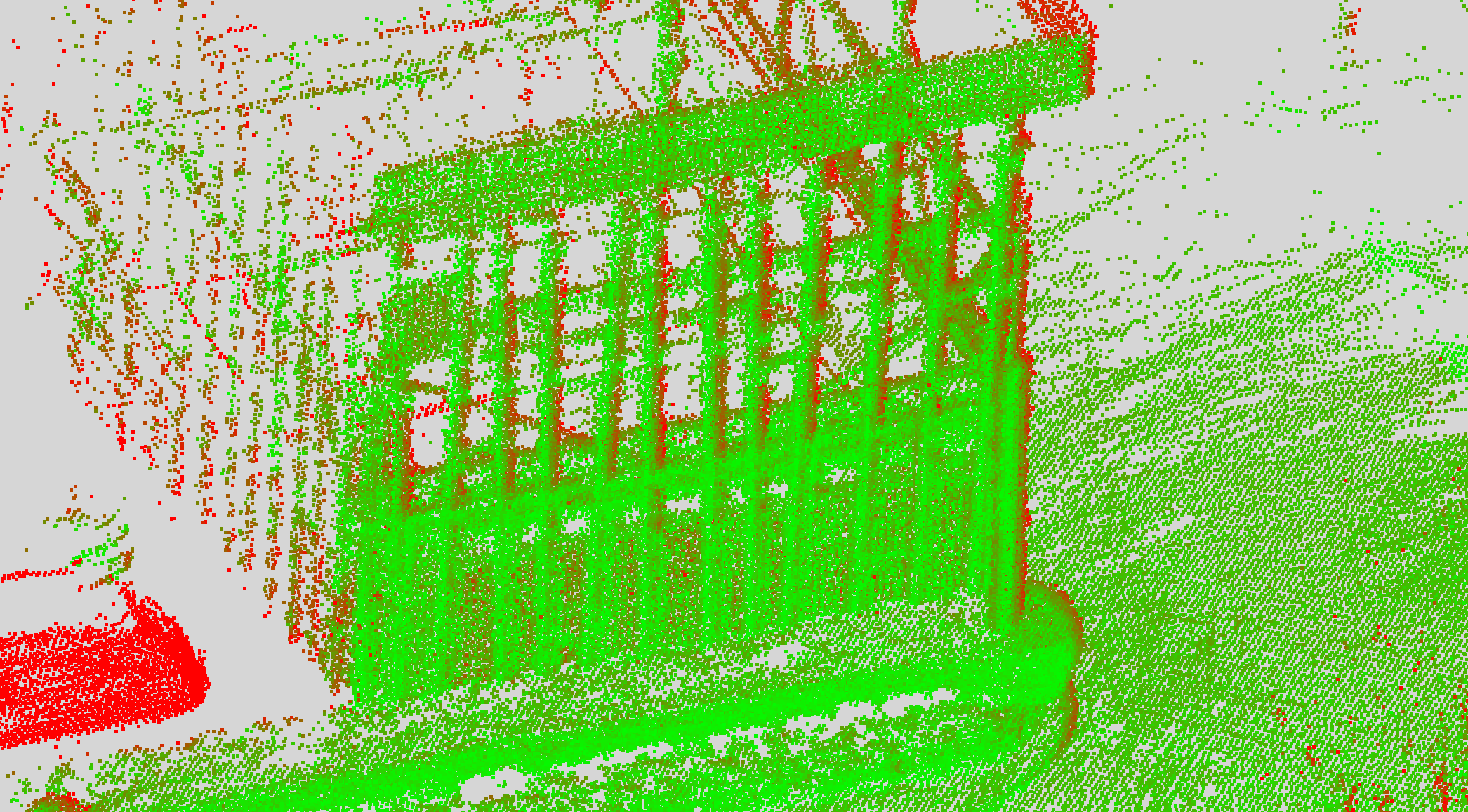}\hfill
\caption{AR Map Detail}
\label{fig:lio_sam_compare_ar_detail}
\end{subfigure}\hfil
\begin{subfigure}{0.5\textwidth}
\includegraphics[width=\textwidth]{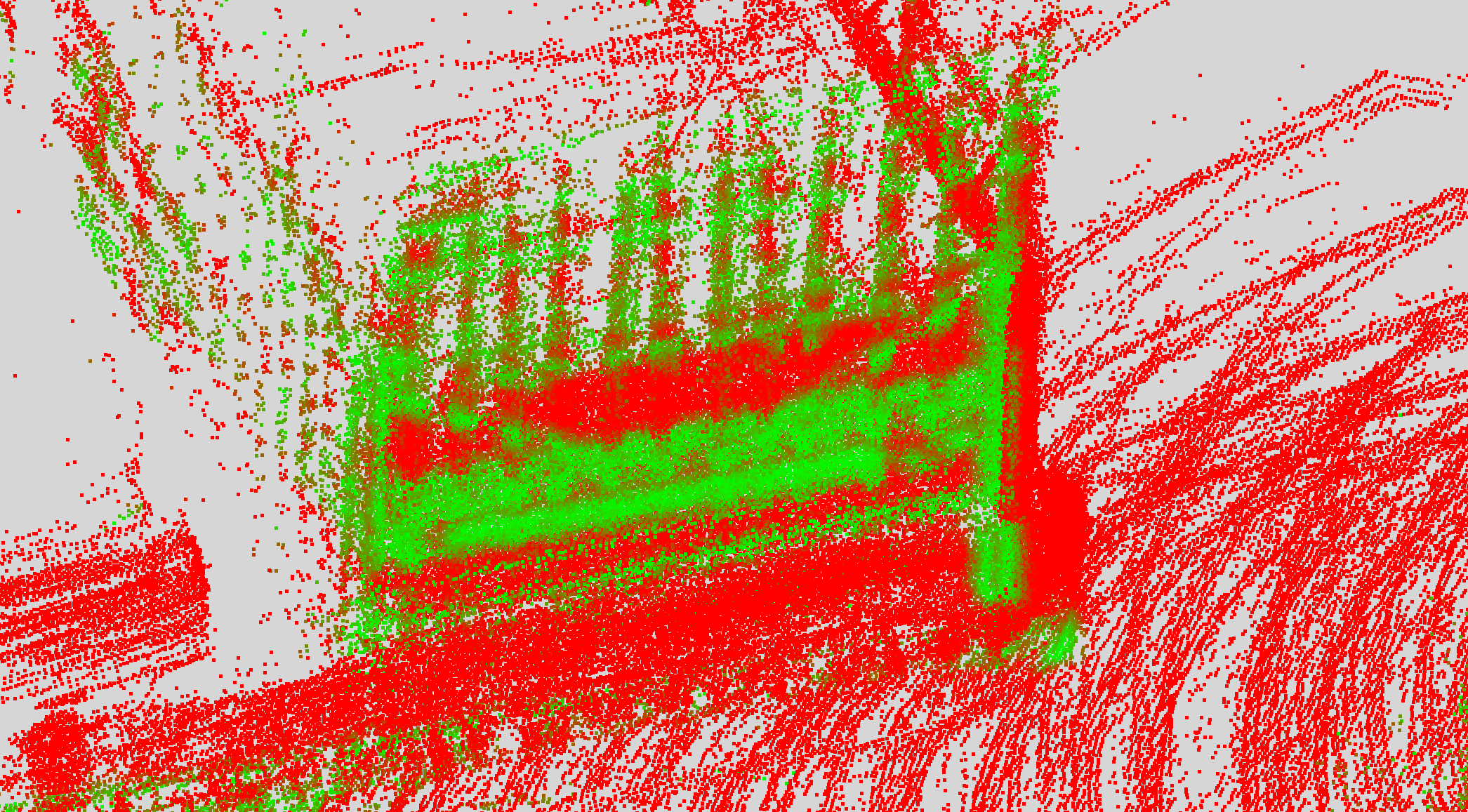}\hfill
\caption{LIO-SAM Detail}
\label{fig:lio_sam_compare_lio_detail}
\end{subfigure}\hfil

\caption{(a) The high-accuracy map from Leica BLK360, (b) Map from AR Mapping, (c) and (d) show a visual comparison between map quality of AR Mapping system and LIO-SAM.}
\label{fig:lio_sam_compare}
\end{figure}

We further compare our results with the state-of-the-art lidar mapping system LIO-SAM~\cite{shan2020lio}. Since the implementation of LIO-SAM only works with single lidar, we run the lidar odometry using only the horizontal lidar for LIO-SAM and AR Mapping. Fig.~\ref{fig:lio_sam_compare_leica} shows the ground truth point cloud map of campus outdoor scene reconstructed by Leica BLK360. Fig.~\ref{fig:lio_sam_compare_backpack} shows the point cloud map reconstructed using AR mapping system. The points are colored based on the error with respect to the Leica Point cloud. The green means the smaller error while the red are large errors. We evaluate the map accuracy with respect to the Leica point cloud. Since some regions may not be covered by the Leica point cloud, we ignore the points with errors larger than $0.2 m$. The mean, median and RMS errors of AR Mapping is $0.077m$, $0.069m$ and $0.091m$ which is smaller than map from LIO-SAM with corresponding error of $0.087m$, $0.080m$ and $0.103m$. Fig.~\ref{fig:lio_sam_compare_ar_detail} and \ref{fig:lio_sam_compare_lio_detail} shows a visual comparison between maps from AR mapping and LIO-SAM respectively. AR map has sharper structures while the LIO-SAM map has relatively blurry details. 


\begin{table}[ht]
    \centering
    \begin{tabular}{c|c|c|c}
    \hline
         Dataset & Mean Err. & Median Err. & RMS Err.\\
        \hline
         Office & 0.046 m & 0.042 m & 0.051 m\\
         \hline
         Campus Building & 0.048 m & 0.037 m & 0.062 m\\
         \hline
         Campus Outdoor & 0.068 m & 0.059 m & 0.082 m\\
         \hline
    \end{tabular}
    \caption{Map accuracy compared with ground truth.}
    \label{tab:lidar_map_accuracy}
\end{table}

\begin{table}[ht]
    \centering
    \begin{tabular}{c|c|c|c}
    \hline
         Dataset & Area & Backpack & Leica\\
        \hline
         Office & 300 $m^2$ & 7 min & 66 min\\
         \hline
         Campus Building & 5000 $m^2$  & 41 min & 120 min\\
         \hline
         Campus Outdoor & 6000 $m^2$ & 33 min & 60 min\\
         \hline
    \end{tabular}
    \caption{Mapping efficiency compared with Leica BLK 360 Stationary laser scanner.}
    \label{tab:lidar_map_efficiency}
\end{table}

\subsection{Evaluation on Pose and Depth Map Accuracy}
\begin{figure}[!t]
\includegraphics[width=0.7\textwidth]{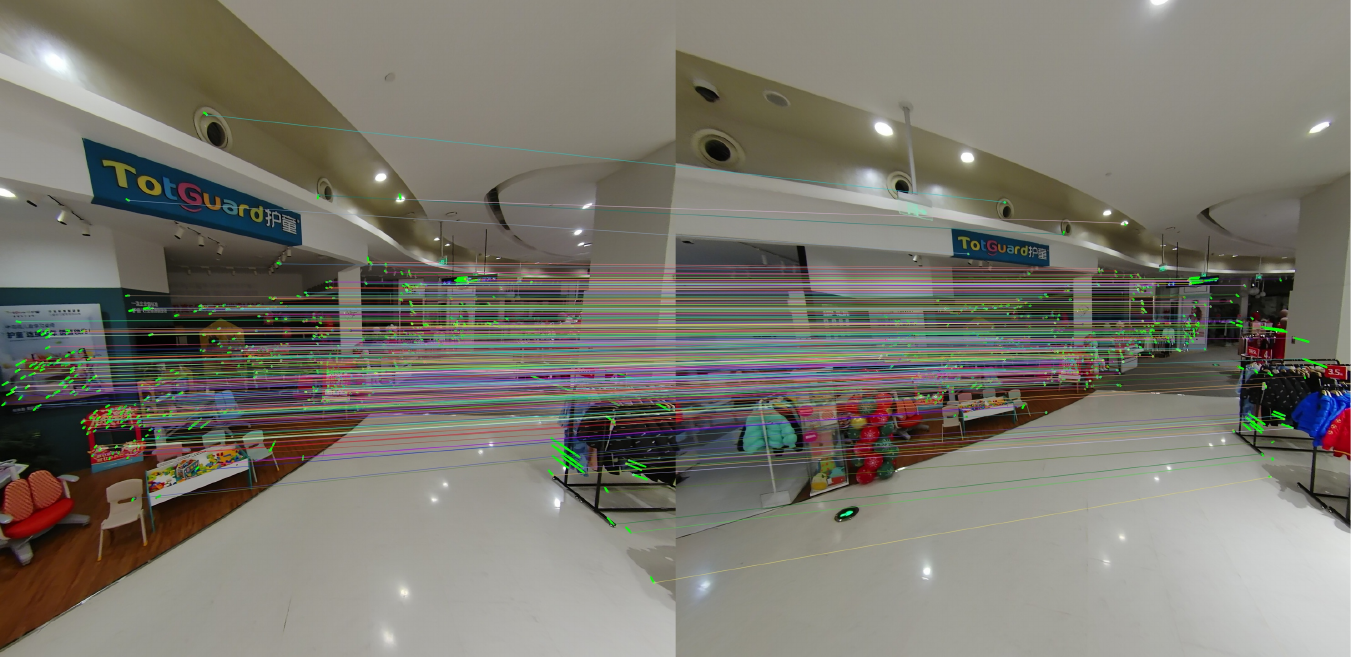}
\caption{Example color image pairs with feature matches captured in WestCity.}
\label{fig:feature_matches}
\end{figure}

\begin{figure}[!t]
\begin{subfigure}{0.45\textwidth}
\includegraphics[width=\textwidth]{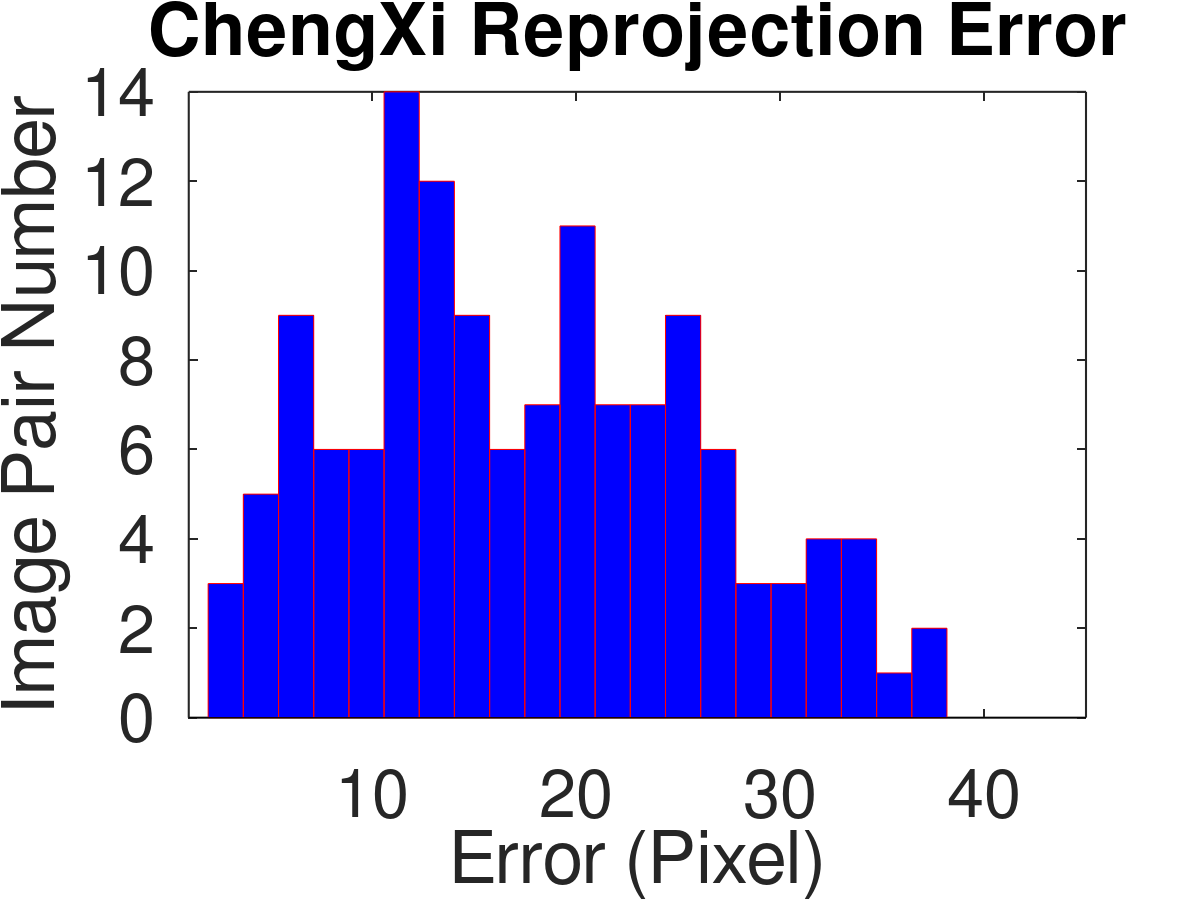}
\caption{}
\end{subfigure}
\begin{subfigure}{0.45\textwidth}
\includegraphics[width=\textwidth]{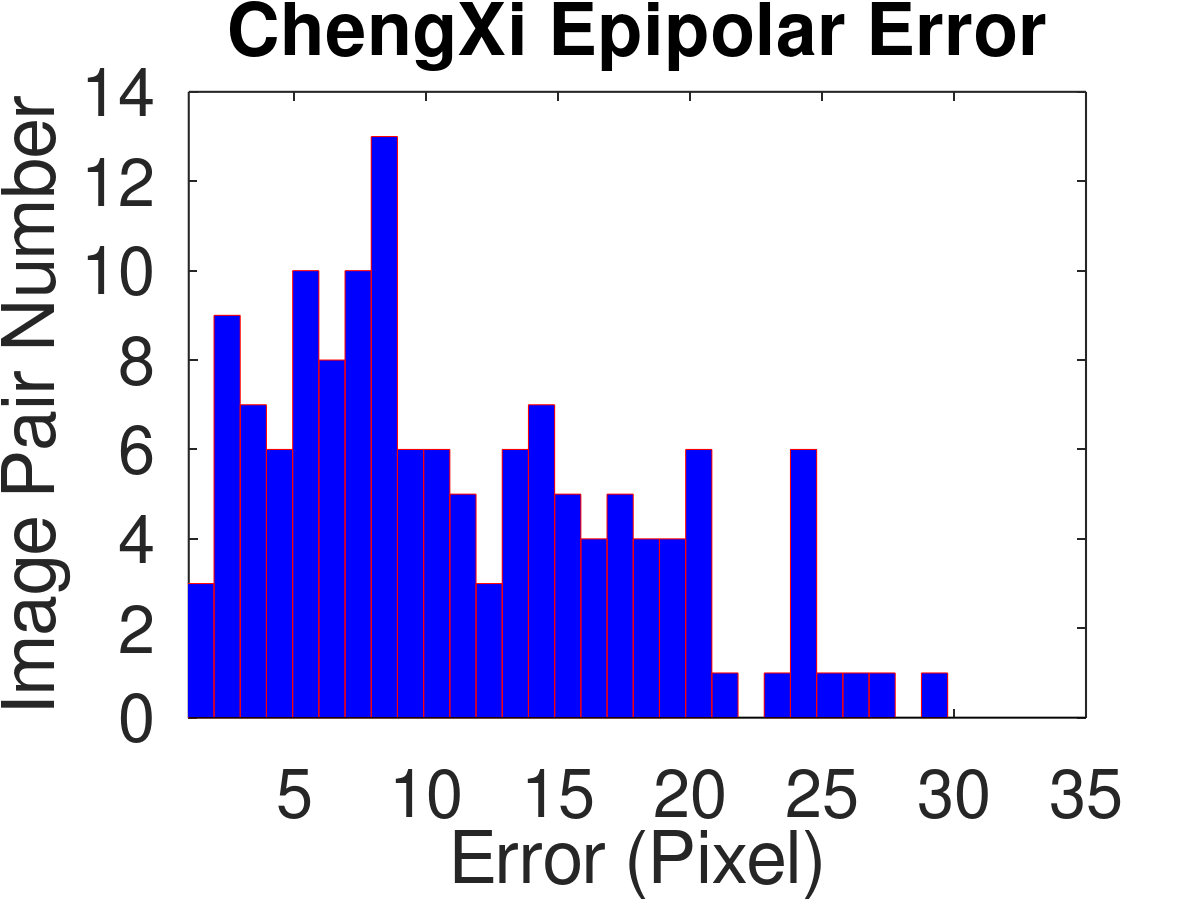}
\caption{}
\end{subfigure}

\begin{subfigure}{0.45\textwidth}
\includegraphics[width=\textwidth]{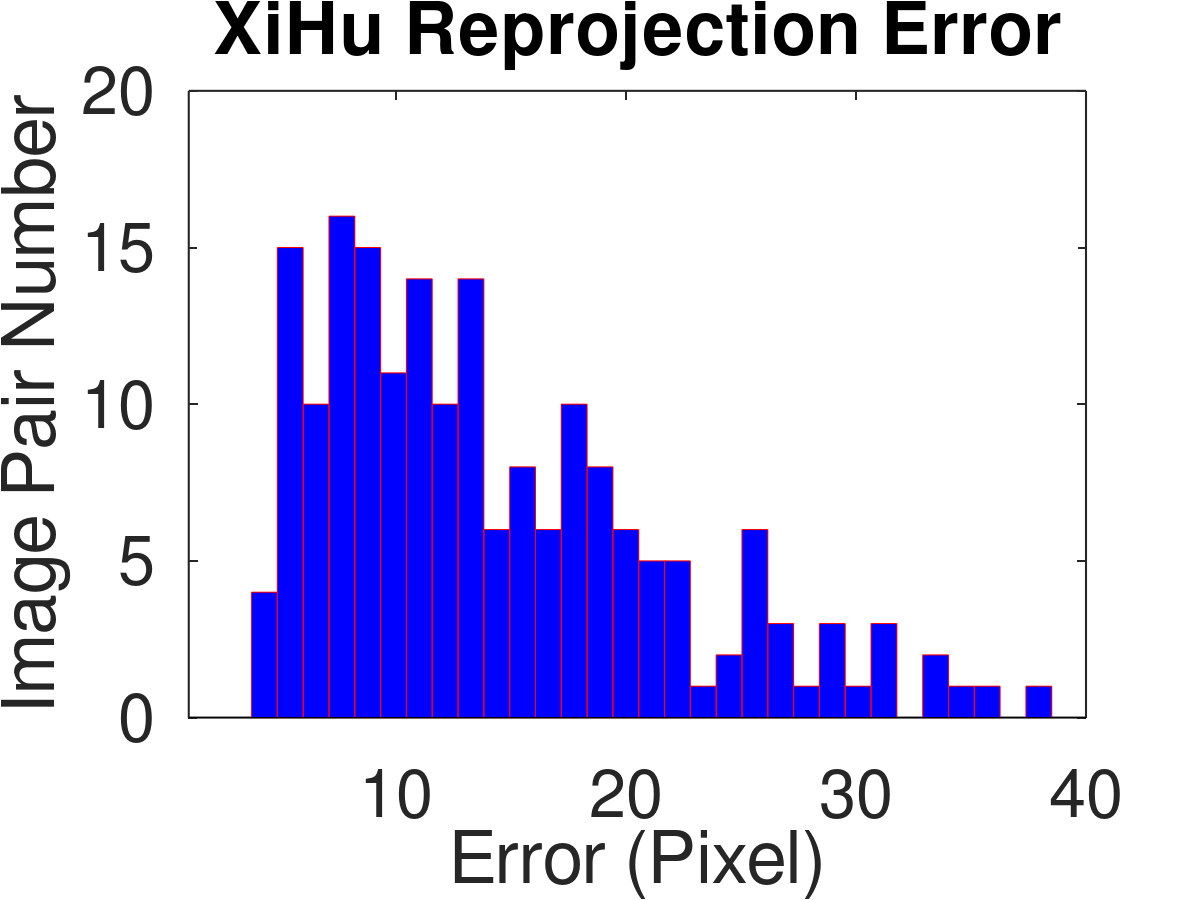}
\caption{}
\end{subfigure}
\begin{subfigure}{0.45\textwidth}
\includegraphics[width=\textwidth]{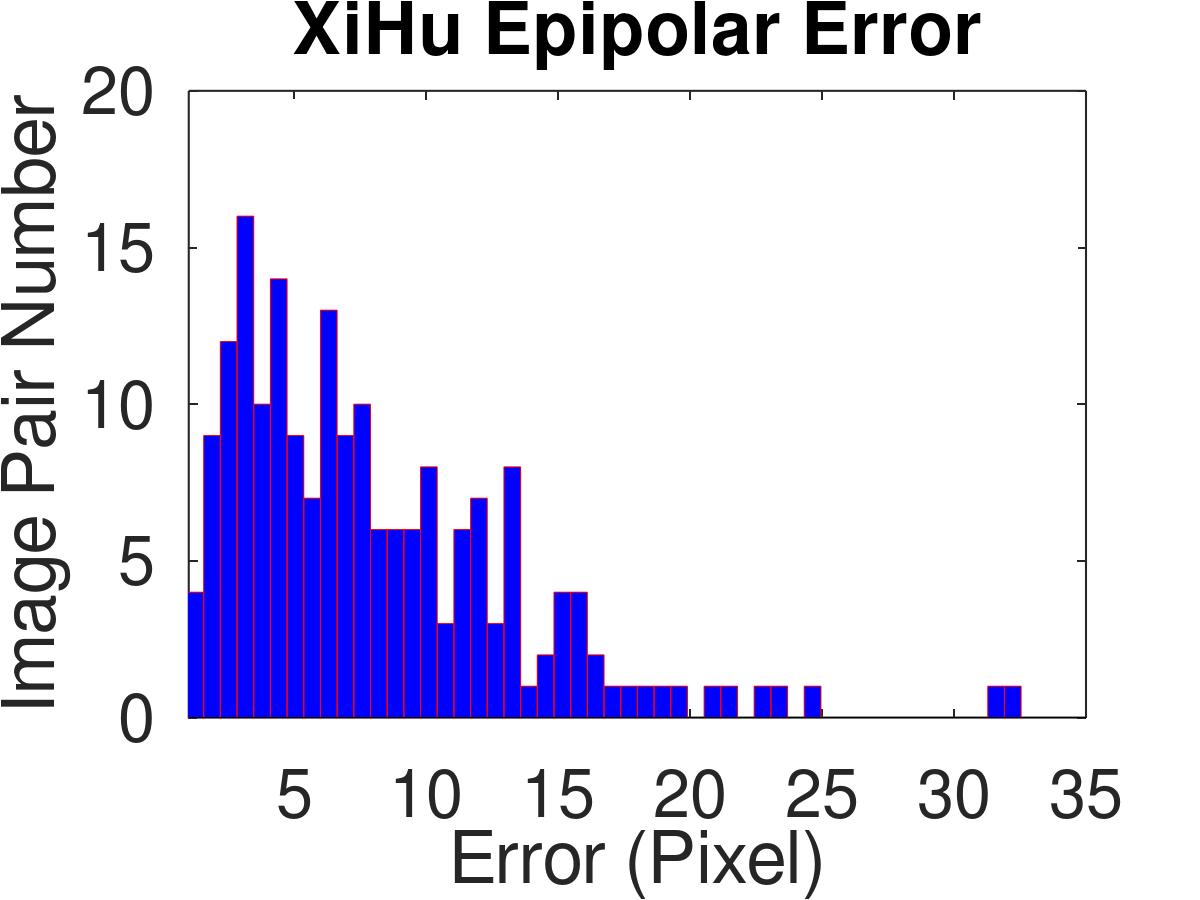}
\caption{}
\end{subfigure}
\caption{The histogram of reprojection error and epipolar error in WestCity and WestLake Datasets.}
\label{fig:error_histogram}
\end{figure}

We further evaluate the accuracy of 6-DOF poses and depth maps on AR Maps captured in two large scale shopping mall, WestCity and WestLake. Each shopping mall has 5 floors and the area of each floor is around 2000 $m^2$. This evaluation is based on the image pairs captured in sequence. As illistrated in Fig.~\ref{fig:feature_matches}, for a specific image pair $(I_i,I_j)$, we first use guided-matching to find high-accuracy SIFT~\cite{lowe1999object} feature matches between them. Since $(I_i,I_{i+1})$ has corresponding depth maps $(D_i,D_j)$ and absolute poses $(T_i,T_j)$ from the AR map, we can project a feature $f^i$ in $I_i$ to $I_j$ and compute the reprojection error and epipolar error as follows,
\begin{equation}
\label{equ:evaluation_reproj}
    \hat{f}^j = \Theta(\Theta^{-1}(f^i, D_i, T_i), T_j)
\end{equation}

\begin{equation}
\label{equ:evaluation_reproj}
    e_1 = \hat{f}^j - f^j, e_2 = Dist(\hat{f}^j, l^j)
\end{equation}
where $\Theta$ is denoted as the projection function from a 3D point to a 2D coordinates in a image plane with the camera' s extrinsic and intrinsic parameters. $\Theta^{-1}$ is the inverse projection from 2D to 3D. $\hat{f}^j$ is the reprojected feature coordinates on $I_j$. $l_j$ is the epipolar line corresponding to $f_i$ in $I_j$. We denote $e_1$ the reprojection error and $e_2$ the epiploar error. The error histograms of AR Map in WestCity and WestLake are presented in Fig.~\ref{fig:error_histogram}. The quantitative evaluation results in terms of the mean and media of $e_1$ and $e_2$ are presented in Table~\ref{tab:error_quantitative}. We also divide the errors by the camera's focal length. This normalized error basically means an angular error regardless of the image resolution. It shows that the $e_1$ and $e_2$ are mostly distributed around $10$ pixels with the color image resolution $2304$ x $3840$ and the normalized error is below $1.5^\circ$.

\begin{table}[ht!]
    \centering
    \begin{tabular}{c|p{1.3cm}|p{1.2cm}|p{1.3cm}|p{1.2cm}}
    \hline
         Dataset & Mean $\bar{e}_1$ & Med. $\Tilde{e}_1$ & Mean $\bar{e}_2$ & Med. $\Tilde{e}_2$\\
        \hline
         WestLake & 7.77px. $0.48^\circ$ & 5.98px. $0.37^\circ$& 23.79px. $1.48^\circ$ & 12.76px. $0.80^\circ$\\
         \hline
         WestCity & 11.13px. $0.69^\circ$ & 8.99px. $0.56^\circ$ & 24.14px. $1.50^\circ$ & 16.86px. $1.05^\circ$ \\
         \hline
    \end{tabular}
    \caption{Quantitative evaluation on $e_1$ and $e_2$ in datasets WestCity and WestLake. In each cell, we show the euclidean pixel error in the first row and the normalized angular error in the second row.}
    \label{tab:error_quantitative}
\end{table}

\subsection{Application of ARMap in Localization}
\begin{figure}[t!]  
\begin{subfigure}{\textwidth}
\includegraphics[width=\linewidth]{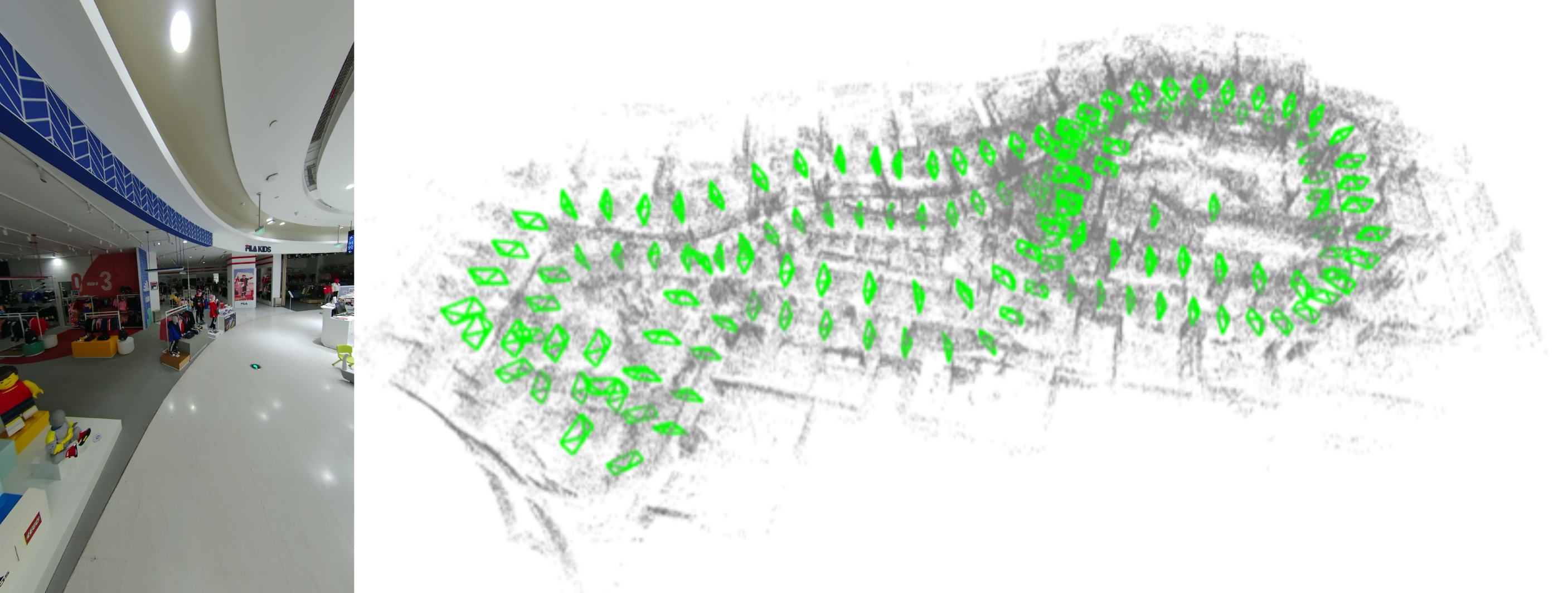}
\caption{}
\label{fig:chengxi_f1-4_map_cams}
\end{subfigure}

\begin{subfigure}{\textwidth}
\includegraphics[width=\linewidth]{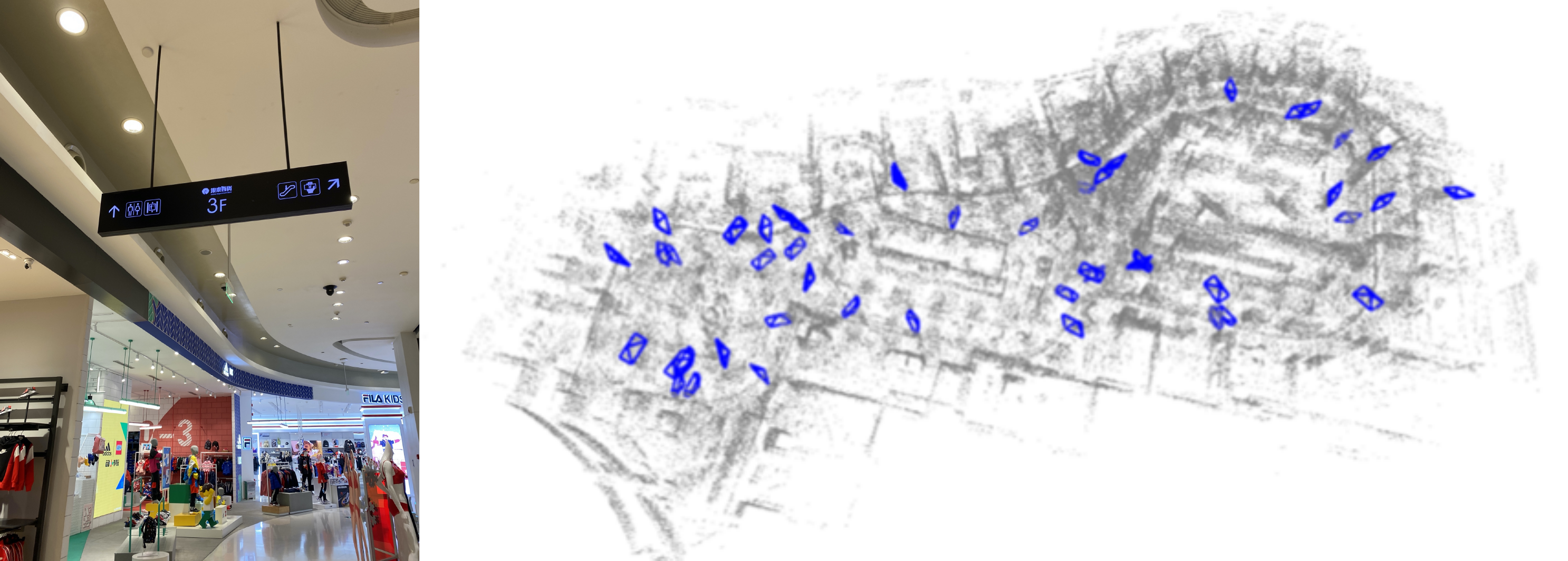}
\caption{}
\label{fig:chengxi_f1-4_query_cams}
\end{subfigure}
\caption{The example and camera poses of (a) scene images  and (b) query images in AR Map of WestCity mall.} 
\label{fig:localization_demo}
\end{figure}

We further verify the effectiveness of AR Map by applying it to a localization task as in real AR applications. We capture a set of query images with resolution $3024 \times 4032$ pixels by a iPhone 11 mobile phone. For each query image, we first use deep image retrieval~\cite{gordo2017end} to search for the scene images in AR Map and then manually label feature correspondences between the query and scene images. Finally, with the known depth information, the 6-DOF pose of query image can be easily computed by solving a 3D-2D PnP problem~\cite{li2012robust}. As shown in Fig.~\ref{fig:localization_demo}, the example query and retrieved scene images are presented along with their poses in the map. We treat those query images successfully localized if the number of inliers is larger than 8 from solving PnP. There are 97 out of 110 query images ($88.2\%$) which can be localized and it shows the AR Map is able to support localization task in the large scene.

\section{conclusion}
In this paper, we propose an end-to-end framework to build and evaluate AR Maps. A backpack scanning system is designed with a unified calibration approach for efficient data capture and the raw data is further processed by a AR Mapping system to generate accurate AR Maps. The feature filtering strategy and submap-based global optimization module ensure accurate trajectory estimation. The stable mapping component is able to fuse the lidar scans to produce high quality AR map even in highly dynamic environment. Finally, we present an approach for systematic evaluation on AR Maps.

\bibliographystyle{IEEEtran}
\bibliography{egbib}

\end{document}